\documentclass[times,referee,twocolumn,final,authoryear]{elsarticle}

\usepackage{graphicx}
\usepackage{subcaption}
\usepackage{float}
\usepackage[justification=justified]{caption}	    %
\usepackage{lscape}                                         %
\usepackage{wrapfig}

\usepackage[lined,ruled,linesnumbered]{algorithm2e}
\usepackage{animate} %

\usepackage{booktabs}                   %
\usepackage{multirow}

\usepackage{makecell}

\usepackage{paralist}
\usepackage{enumitem}

\usepackage{bm}                          %
\usepackage{epsfig}                      %
\usepackage{graphicx}                  %
\usepackage{times}
\usepackage{mathtools}
\usepackage{amssymb,amsmath}   %

\usepackage{units}
\usepackage{color}

\usepackage{comment}

\usepackage{url}  %
\usepackage[breaklinks=true,colorlinks,urlcolor=blue,citecolor=blue,linkcolor=blue,bookmarks=false]{hyperref}
\usepackage{xspace}
\usepackage[table]{xcolor}

\usepackage{nicefrac}
\usepackage{microtype}
\usepackage{setspace}
\usepackage{grfext}
\PrependGraphicsExtensions*{.jpg,.png,.PNG}

\def\eg{e.g.,~}               %
\def\ie{i.e.,~}               %
\def\etc{etc}                 %

\newlength\paramargin
\newlength\figmargin

\newlength\secmargin
\newlength\figcapmargin
\newlength\tabcapmargin

\newcommand{\red}{\textcolor{red}}
\newcommand{\blue}{\textcolor{blue}}
\newcommand{\green}{\textcolor{green}}

\newcommand{\mpage}[2]
{
\begin{minipage}{#1\linewidth}\centering
#2
\end{minipage}
}

\newcommand{\topic}[1]
{
\vspace{1mm}\noindent\textbf{#1}
}

\newcommand{\figcaption}[2]
{
\caption{
\textbf{#1.}  %
#2            %
}
}

\newcommand{\secref}[1]{Section~\ref{sec:#1}}
\newcommand{\figref}[1]{Figure~\ref{fig:#1}} 
\newcommand{\tabref}[1]{Table~\ref{tab:#1}}

\newcommand{\algref}[1]{Algorithm~\ref{#1}}

\long\def\ignorethis#1{}

\newcommand{\tb}[1]{\textbf{#1}}

\newbox\jsavebox%

\newcommand{\best}[1]{\red{\textbf{#1}}}
\newcommand{\second}[1]{\blue{\underline{#1}}}

\newif\ifmarked
\markedfalse     %

\def\xi{\mathbf{x}_i}

\graphicspath{{figure}, {example}}

\usepackage{ycviu}
\usepackage{framed,multirow}

\usepackage{latexsym}
\usepackage{animate}

\definecolor{newcolor}{rgb}{.8,.349,.1}

\newcommand{\heading}[1]{\noindent\textbf{#1}}
\journal{Computer Vision and Image Understanding}

\begin{document}

\ifpreprint
  \setcounter{page}{1}
\else
  \setcounter{page}{1}
\fi

\begin{frontmatter}

\title{Learning Representational Invariances for Data-Efficient Action Recognition}

\author[1]{Yuliang \snm{Zou}} 
\author[2]{Jinwoo \snm{Choi}\corref{cor1}}
\cortext[cor1]{Corresponding author: 
  Tel.: +82-31-201-3758.
}
\ead{jinwoochoi@khu.ac.kr}
\author[3]{Qitong \snm{Wang}}
\author[4]{Jia-Bin \snm{Huang}}

\address[1]{Department of Electrical and Computer Engineering, Virginia Tech, VA, USA}
\address[2]{Department of Computer Science and Engineering, Kyung Hee University, Yongin, Korea}
\address[3]{Department of Computer and Information Sciences, University of Delaware, DE, USA}
\address[4]{Department of Computer Science, University of Maryland College Park, MD, USA}

\received{1 May 2013}
\finalform{10 May 2013}
\accepted{13 May 2013}
\availableonline{15 May 2013}
\communicated{S. Sarkar}

\begin{abstract}

Data augmentation is a ubiquitous technique for improving image classification when labeled data is scarce.
Constraining the model predictions to be invariant to diverse data augmentations effectively injects the desired representational invariances to the model (e.g., invariance to photometric variations) and helps improve accuracy.
Compared to image data, the appearance variations in videos are far more complex due to the additional temporal dimension.
Yet, data augmentation methods for videos remain under-explored.
This paper investigates various data augmentation strategies that capture different video invariances, including photometric, geometric, temporal, and actor/scene augmentations.
When integrated with existing semi-supervised learning frameworks, we show that our data augmentation strategy leads to promising performance on the Kinetics-100/400, Mini-Something-v2, UCF-101, and HMDB-51 datasets in the low-label regime.
We also validate our data augmentation strategy in the fully supervised setting and demonstrate improved performance.

\end{abstract}

\begin{keyword}
\MSC 41A05\sep 41A10\sep 65D05\sep 65D17
\KWD 3D human pose and shape estimation \sep Self-supervised learning \sep Occlusion handling

\end{keyword}

\end{frontmatter}

\section{Introduction}
\label{sec:intro}

Deep neural networks have shown rapid progress in video action recognition~\citep{Simonyan-NeurIPS-2014,Carreira-CVPR-2017,Tran-CVPR-2018,Xie-ECCV-2018,feichtenhofer2018slowfast,lin2019tsm,feichtenhofer2020x3d,yang2020temporal}.
However, these approaches rely on training a model on a massive amount of \emph{labeled} videos. 
For example, the SlowFast networks~\citep{feichtenhofer2018slowfast}, R(2+1)D~\citep{Tran-CVPR-2018}, and I3D~\citep{Carreira-CVPR-2017} are pre-trained on the Kinetics-400 dataset~\citep{kay2017kinetics}, containing $300K\sim650K$ \emph{manually labeled} and \emph{temporally trimmed} videos. 
The dependency on large-scale annotated video datasets is not scalable because manual labeling of videos is expensive, time-consuming, and error-prone. 
Hence, it is of great interest to investigate new approaches to improve data efficiency.

\begin{figure*}[ht]
\centering

\includegraphics[width=0.24\linewidth]{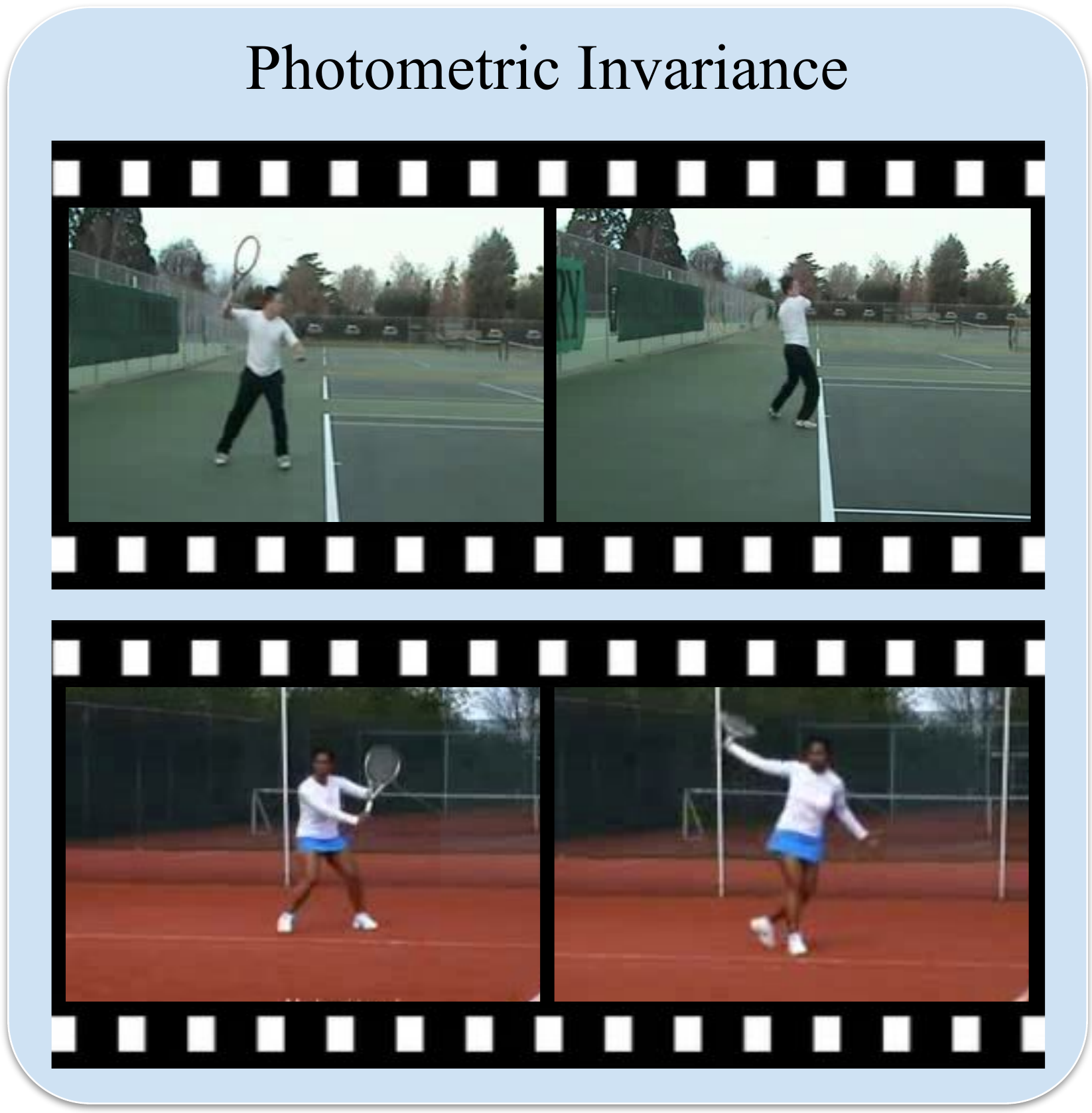}\hfill
\includegraphics[width=0.24\linewidth]{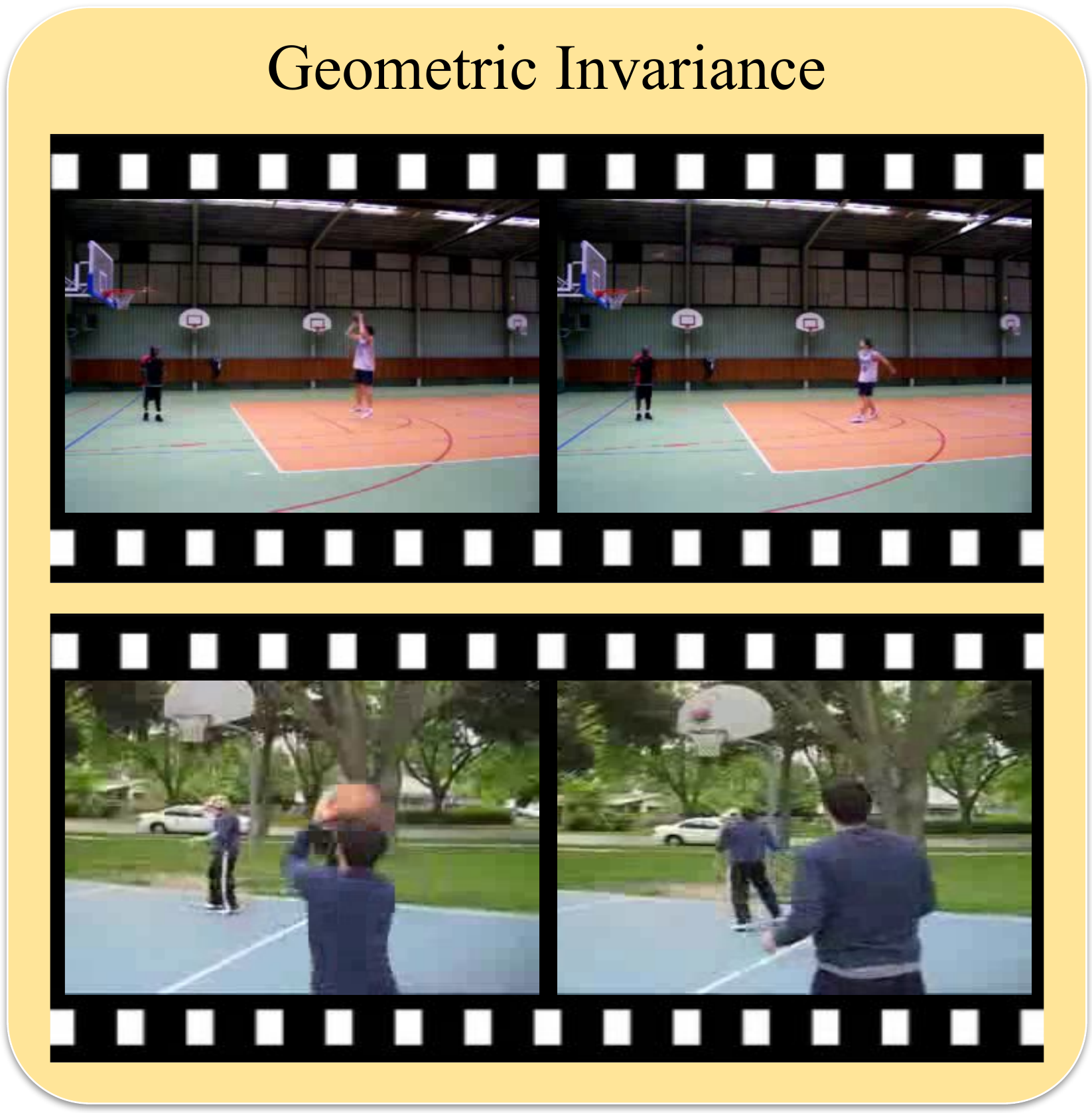}\hfill
\includegraphics[width=0.24\linewidth]{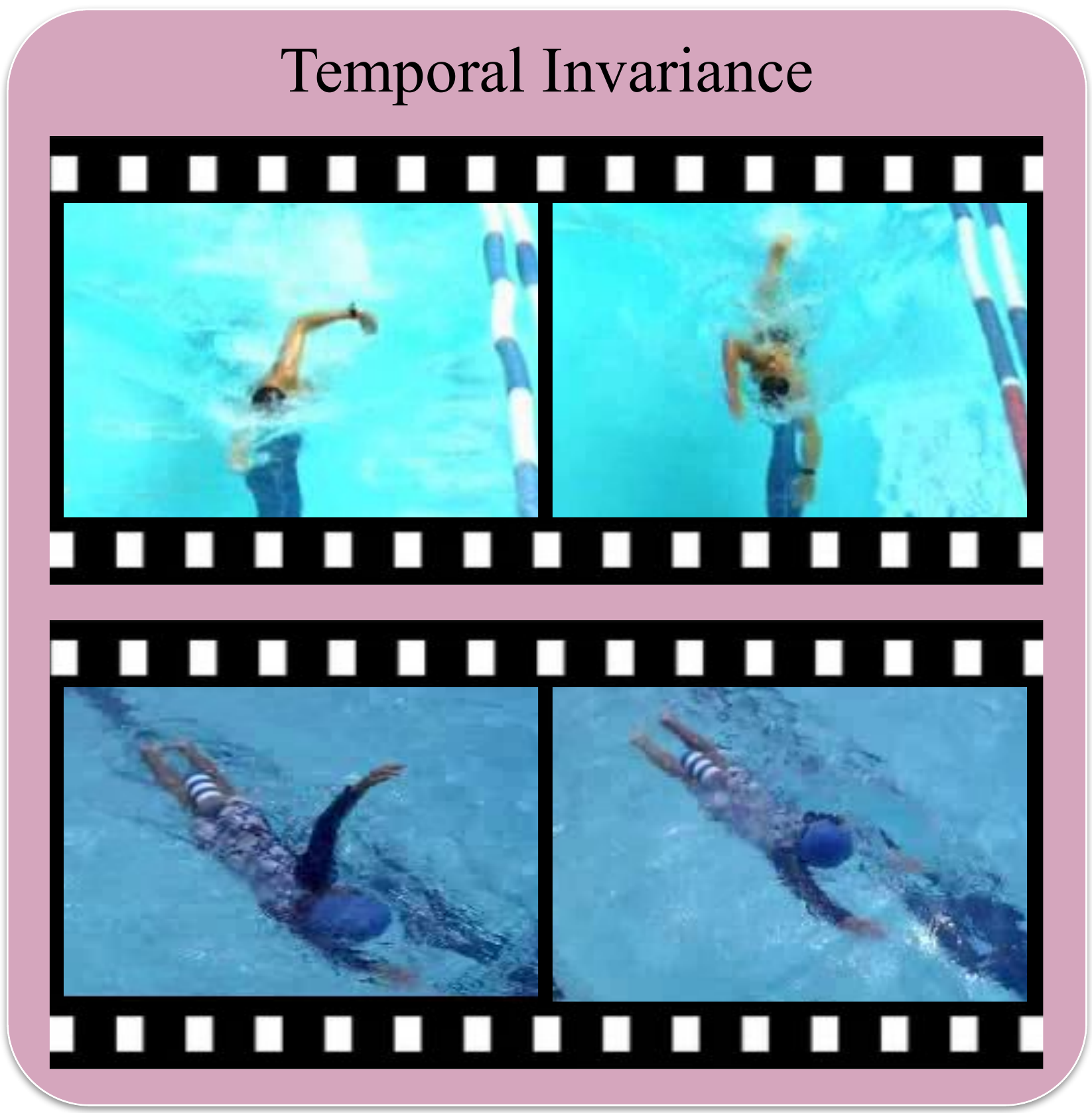}\hfill
\includegraphics[width=0.24\linewidth]{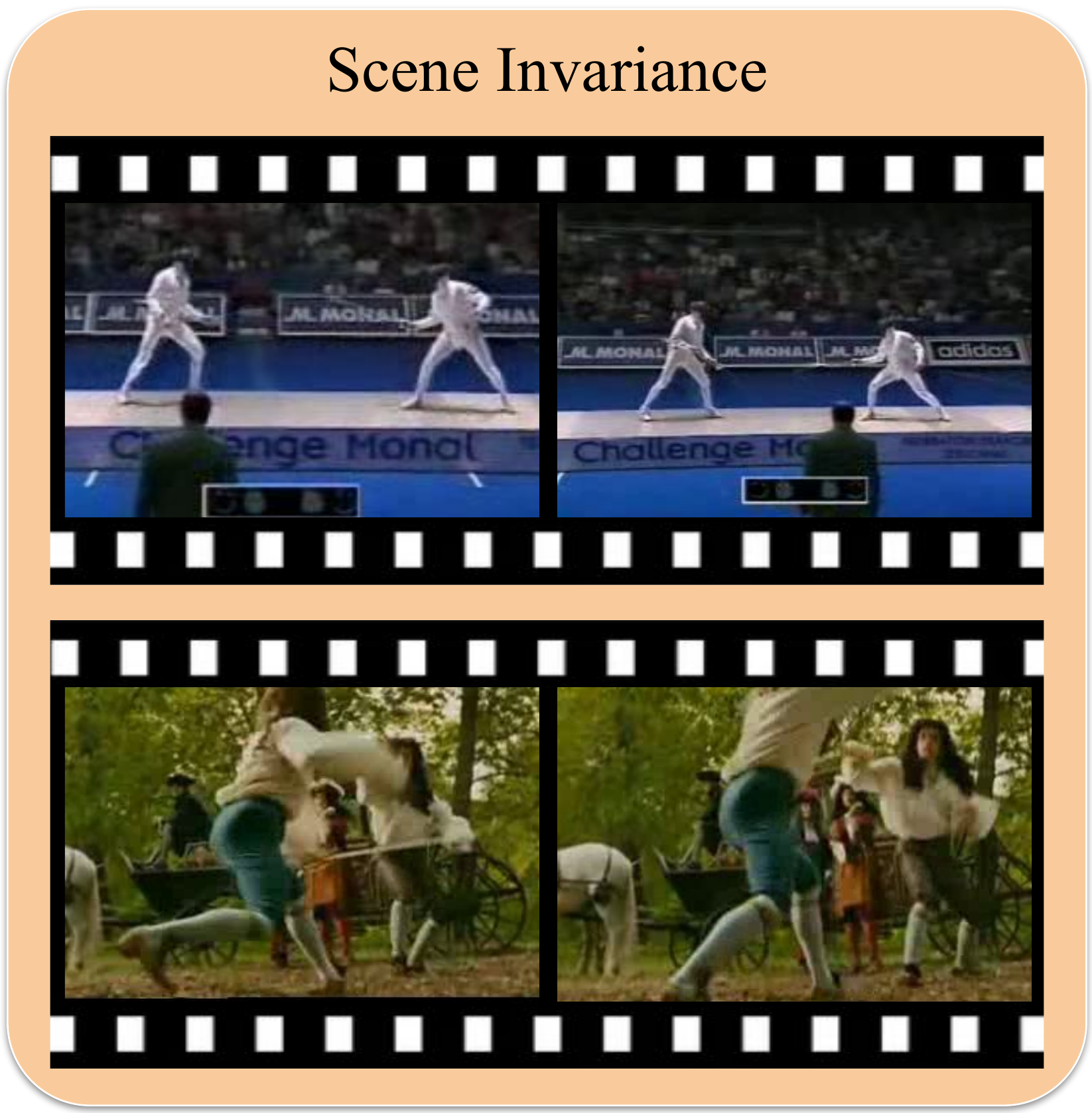}\hfill

\figcaption{Representational invariances for video action recognition}
{
Humans recognize actions in videos effortlessly, even in the presence of large photometric, geometric, temporal, and scene variations. 
Injecting these task-specific video invariances through data augmentations could improve data efficiency.
For each shaded block, we show two example videos of the \emph{same} action, \ie ``Tennis swing'', ``Basketball'', ``Front Crawl'', and ``Fencing''.
}
\label{fig:motivation}
\end{figure*}

Data augmentation is a simple yet effective approach for improving data efficiency.
Early work uses simple/weak augmentations to generate additional \emph{realistic-looking} samples by applying geometric transformations (\eg random scaling and cropping).
Such approaches have been widely used in training supervised image classification models~\citep{krizhevsky2012imagenet,he2016deep,huang2017densely}.
While many realistic-looking visual examples are generated, these \emph{weak} augmentations are ineffective in increasing the training data diversity because the augmented samples remain highly similar to the original ones.
Recently, \emph{diverse/strong} augmentations have been proposed~\citep{cubuk2018autoaugment,cubuk2020randaugment,lim2019fast}.
By applying aggressive photometric and geometric transformations to the original data, strong augmentations help generate diverse training data and improve model performance in the supervised learning setting.
Later, strong data augmentations have also shown their potential in consistency-based semi-supervised learning frameworks~\citep{sohn2020fixmatch,xie2019unsupervised}.
The intuition behind the success of strong data augmentations in supervised and semi-supervised settings lies in learning \emph{representational invariances} via regularizing a model to generate consistent predictions for diverse augmented views of the same data.

In this paper, we investigate and study \emph{strong} data augmentation for video.
While strong data augmentation in the image domain has been extensively studied~\citep{devries2017improved,cubuk2018autoaugment,lim2019fast,yun2019cutmix,cubuk2020randaugment}, data augmentation techniques for videos remain under-explored.
Recent self-supervised video representation learning approaches~\citep{han2019video,han2020memory,tschannen2019self,xu2019self,wang2020self,qian2020spatiotemporal} show the benefits of leverage some specific video invariances (\ie spatial, temporal, etc.). 
However, these approaches only utilize partial aspects of the desired representation invariances.
In contrast, we provide a comprehensive study in semi-supervised and fully-supervised settings.
As shown in \figref{motivation}, we explore the following invariances: \emph{photometric (color)}, \emph{geometric (spatial)}, \emph{temporal}, \emph{scene (background)}, and investigate the corresponding data augmentation strategies that encourage the model learning these invariances.

Our results show that all these augmentation techniques help improve performance. 
With all the augmentations (as shown in \figref{overview}), we achieve favorable results on the Kinetics-100~\citep{jing2020videossl}, UCF-101~\citep{soomro2012ucf101}, and HMDB-51~\citep{kuehne2011hmdb} datasets in both low-label and full-label settings.

To summarize, we make the following contributions.
\begin{itemize}
\item We study various strong video data augmentation strategies for data-efficient action recognition.
\item We introduce a novel scene augmentation strategy, ActorCutMix, to encourage scene invariance, which is crucial for action recognition.
\item Our strong data augmentation strategies significantly improve the data efficiency in both the semi-supervised and fully-supervised learning settings, achieving promising results in standard video action recognition benchmarks.
\item Our source code and pre-trained models are publicly available at \url{https://github.com/vt-vl-lab/video-data-aug}.
\end{itemize}

\section{Related Work}
\label{sec:related}

\topic{Video recognition models.}
Recent advances in video recognition focus on improving the network architecture design (\eg two-stream networks~\citep{Simonyan-NeurIPS-2014,Feichtenhofer-CVPR-2016}, 
3D CNNs~\citep{Tran-ICCV-2015,Carreira-CVPR-2017,hara3dcnns,Wang-CVPR-2018}, 
2D and 1D separable CNNs~\citep{Tran-CVPR-2018,Xie-ECCV-2018}, 
\ifmarked
\red{video Transformers~\citep{arnab2021vivit,fan2021multiscale,liu2022video},}
\else
video Transformers~\citep{arnab2021vivit,fan2021multiscale,liu2022video},
\fi
incorporating long-term temporal contexts~\citep{Wang-CVPR-2018,feichtenhofer2018slowfast,wu2019long}), and training efficiency~\citep{wu2020multigrid}.
Our focus in this work lies in improving \emph{data efficiency} of video action classification by exploring data augmentation strategies from various perspectives: photometric, geometric, temporal, and actor/scene.

\topic{Data augmentation.}
Data augmentation is an essential component in modern deep neural network training.
Early approaches~\citep{laine2016temporal,sajjadi2016regularization} only apply weak augmentations such as random translation and cropping.
\ifmarked
\red{Depending on the downstream tasks, various invariances are needed to improve the performance. For instance, random Gaussian, Dropout noise~\citep{bachman2014learning} and adversarial noise~\citep{miyato2018virtual} have also been proposed for semi-supervised learning, leading to improved performance.
Sometimes, data augmentation techniques are used to encourage models sensitive to variances to certain transformations: rotation~\citep{gidaris2018unsupervised} and time-shift~\citep{patrick2021compositions} in video.}
\else
Depending on the downstream tasks, various invariances are needed to improve the performance. For instance, random Gaussian, Dropout noise~\citep{bachman2014learning} and adversarial noise~\citep{miyato2018virtual} have also been proposed for semi-supervised learning, leading to improved performance.
Sometimes, data augmentation techniques are used to encourage models sensitive to variances to certain transformations: rotation~\citep{gidaris2018unsupervised} and time-shift~\citep{patrick2021compositions} in video.
\fi
Learning-based data augmentation approaches~\citep{cubuk2018autoaugment,lim2019fast} aim to avoid the manual design of data transformations.
Such networks learn to adjust the data augmentation policy according to the feedback on a held-out (labeled) validation set.
In semi-supervised classification,
recent methods~\citep{sohn2020fixmatch,xie2019unsupervised} apply strong image space augmentations (\eg RandAugment~\citep{cubuk2020randaugment}) by cascading color jittering, geometric transformations, and regional dropout~\citep{devries2017improved,yun2019cutmix}, achieving state-of-the-art performance.
Instead of perturbing the unlabeled images in the pixel space, several approaches~\citep{kuo2020featmatch,wang2020regularizing} propose to augment training examples in the feature space to complement conventional image space augmentations, providing further improvement.

Most existing works design the data augmentation strategy specifically for \emph{images}. 
\ifmarked
\red{There are only a few works on data augmentation for video action recognition~\citep{yun2020videomix, zhang2020self, patrick2021compositions}.
Similar to our ActorCutMix augmentation, VideoMix~\citep{yun2020videomix} cuts and pastes a spatio-temporal cube from one video to another video.
However, VideoMix does not consider human regions during the cut and paste operation, while the proposed ActorCutMix explicitly cuts and pastes human regions from one video to another.
In this work, in the context of data-efficient \emph{video} action recognition, we extensively study data augmentation from various perspectives: photometric, geometric, temporal, and actor/scene.}
\else
There are only a few works on data augmentation for video action recognition~\citep{yun2020videomix, zhang2020self, patrick2021compositions}.
Similar to our ActorCutMix augmentation, VideoMix~\citep{yun2020videomix} cuts and pastes a spatio-temporal cube from one video to another video.
However, VideoMix does not consider human regions during the cut and paste operation, while the proposed ActorCutMix explicitly cuts and pastes human regions from one video to another.
In this work, in the context of data-efficient \emph{video} action recognition, we extensively study data augmentation from various perspectives: photometric, geometric, temporal, and actor/scene.
\fi

\topic{Semi-supervised learning.}
Semi-supervised learning (SSL) improves the performance using abundant unlabeled data, alleviating the need for manual annotations.
Most recent SSL approaches adopt either one of the following two strategies: 
(1) consistency regularization~\citep{laine2016temporal,sajjadi2016regularization,tarvainen2017mean,miyato2018virtual,kuo2020featmatch,xie2019unsupervised}, and 
(2) entropy minimization~\citep{grandvalet2005semi,lee2013pseudo}.
The key insight of consistency regularization is that a model should generate \emph{consistent} predictions for the same (unlabeled) data undergone different transformations/perturbations.
Recently, holistic approaches~\citep{berthelot2019mixmatch,berthelot2019remixmatch,sohn2020fixmatch,xie2019unsupervised} that combine both the SSL strategies (consistency and entropy minimization) have been proposed to tackle the semi-supervised image classification task effectively.
The consistency regularization within these SSL frameworks effectively encourages representational invariances to strongly-augmented views.

In the low-label settings, we leverage the FixMatch framework~\citep{sohn2020fixmatch} to validate the efficacy of our video data augmentations.
We also demonstrate improved results when integrating our augmentation strategies with another recent SSL framework, UDA~\citep{xie2019unsupervised}.

A few recent works propose to apply the SSL framework to the video domain~\citep{jing2020videossl,singh2021semi}.
They focus on \emph{algorithmic} improvement for video SSL. 
In contrast, our work complements these recent advances by exploring strong data augmentation strategies for videos.

\topic{Self-supervised learning.}
Self-supervised learning is a technique used to learn representations when external labels are unavailable, \eg unsupervised setting. 
In the self-supervised learning framework, people define pretext tasks that are expected to be useful for learning more generalizable representations. 
In video domain, there have been pretext tasks for frame sorting~\citep{lee2017unsupervised}, clip order verification~\citep{misra2016shuffle} and prediction~\citep{xu2019self}, speed prediction~\citep{epstein2019oops,benaim2020speednet}, and future prediction~\citep{han2019video,han2020memory}.

Recently, contrastive learning has emerged as a powerful tool for learning visual representations~\citep{sermanet2018time,han2019video,chen2020simple,chen2020improved,he2020momentum,purushwalkam2020demystifying,singh2021semi,xiao2020should}.
In contrastive learning, a model learns representations by instance discrimination. 
These methods encourage feature embeddings from different augmentations of the same data to be similar and feature embeddings from different data instances to be dissimilar to each other.
Contrastive learning can be viewed as injecting visual invariances. 
It pulls the representations of different augmented views of the same instance together, enforcing invariances to the selected data augmentations.
\ifmarked
\red{To further demonstrate the effectiveness of our augmentation strategy, we validate its compatibility with contrastive learning.
Instead of injecting visual invariances in pure self-supervised learning, we conduct experiments in a semi-supervised learning setting by replacing the consistency regularization with a contrastive objective, following~\citet{singh2021semi}.}
\else
To further demonstrate the effectiveness of our augmentation strategy, we validate its compatibility with contrastive learning.
Instead of injecting visual invariances in pure self-supervised learning, we conduct experiments in a semi-supervised learning setting by replacing the consistency regularization with a contrastive objective, following~\citet{singh2021semi}.
\fi

\section{Video Data Augmentations}
\label{sec:method}
We explore strong and diverse video data augmentations mainly in the low-label setting.
We formulate low-label video action recognition as a semi-supervised classification problem.

We first describe a consistency-based semi-supervised classification formulation in \secref{semi}.
We then present our \emph{intra-clip} data augmentation strategies (photometric, geometric, temporal) in \secref{intra}.
Next, we propose a \emph{cross-clip} human-centric data augmentation operation, ActorCutMix, in \secref{cross}.
We describe how we combine all these data augmentation operations to construct the final strong data augmentation strategy in \secref{combine}.

\begin{figure}[t]
    \centering
    \includegraphics[width=0.95\linewidth]{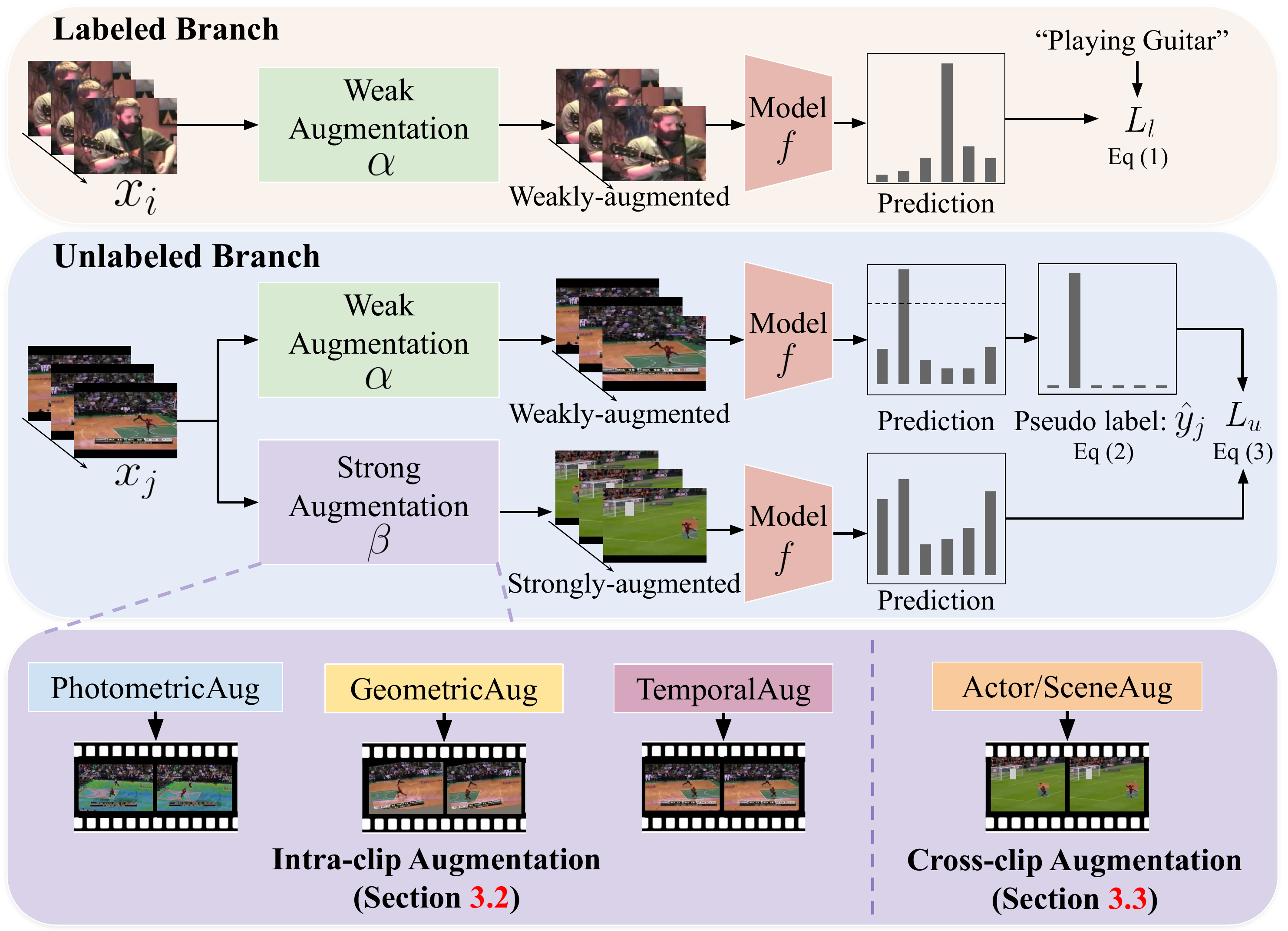}
    \figcaption{Overview of data-efficient action recognition pipeline}{
    In a low-label regime, we adopt the state-of-the-art consistency-based semi-supervised learning framework~\citep{sohn2020fixmatch} to validate the effectiveness of our explored strong data augmentation strategies.
    Our contributions lie in the exploration of the \emph{strong video augmentation} $\beta(\cdot)$.
    }
    \label{fig:overview}
\end{figure}

\subsection{Consistency-based semi-supervised learning}
\label{sec:semi}
Considering a multi-class classification problem, we denote $\mathcal{X} = \{(x_i, y_i)\}_{i=1}^{N_l}$ as the \emph{labeled} training set, where $x_i \in \mathcal{R}^{T\times H\times W\times 3}$ is the $i$-th sampled video clip, $y_i$ is the corresponding one-hot ground truth label, and $N_l$ is the number of data points in the labeled set.
Similarly, we denote $\mathcal{U} = \{x_j\}_{j=1}^{N_u}$ as the \emph{unlabeled} set, where $N_u$ is the number of data points in the unlabeled set.
We use $f_{\theta}$ to denote a classification model with trainable parameters $\theta$.
We use $\alpha(\cdot)$ to represent the weak (standard) augmentation (\ie random horizontal flip, random scaling, and random crop in video action recognition), and $\beta(\cdot)$ to represent the \emph{strong} data augmentation strategies (our focus).

We present an overview of the state-of-the-art semi-supervised classification pipeline~\citep{sohn2020fixmatch} in~\figref{overview}.
We denote an input video clip consists of $T$ frames as $x_i$ throughout the paper.
Given a mini-batch of \emph{labeled} data $\{(x_i, y_i)\}_{i=1}^{B_l}$, we minimize the standard cross-entropy loss $L_l$ defined as
\begin{align}
    L_l = -\frac{1}{B_l} \sum_{i=1}^{B_l} 
    y_i \log f_{\theta}(\alpha(x_i)).
    \label{eq:loss_l}
\end{align}

For a mini-batch of \emph{unlabeled} data $\{x_j\}_{j=1}^{B_u}$, we enforce the model prediction consistency.
More specifically, we generate pseudo-label $\hat{y}$ for the unlabeled data via confidence thresholding
\begin{align}
    \mathcal{C} = \{x_j | \text{max} f_{\theta}(\alpha(x_j)) \geq \tau\},
\end{align}
where $\tau$ denotes a pre-defined threshold and $\mathcal{C}$ is the confident example set for the current mini-batch.
We then convert the confident model predictions $f_{\theta}(\alpha(x_j))$ to one-hot labels $\hat{y}_j$ by taking \textit{argmax} operation.
We optimize a cross-entropy loss $L_u$ for the confident set of unlabeled examples.
\begin{align}
    L_u = -\frac{1}{B_u} \sum_{x_j \in \mathcal{C}}
    \hat{y}_j \log f_{\theta}(\beta(x_j)).
    \label{eq:loss_u}
\end{align}

Our overall training objective is the summation of \eqref{eq:loss_l} and \eqref{eq:loss_u}.
\begin{align}
    L = L_l + \lambda_u L_u.
\end{align}
We set $\lambda_u=1$ and find it performs well empirically.

\ifmarked
\red{
Note that it is easy to switch from a consistency-based semi-supervised learning framework to a contrastive-learining-based semi-supervised learning formulation, by replacing the consistency regularization in \eqref{eq:loss_u} to a contrastive objective.
In this paper, we use the temporal contrastive objective proposed in TCL~\citep{singh2021semi} to further demonstrate the effectiveness of our video data augmentations. 
We refer to \citet{singh2021semi} for more details.
}
\else
Note that it is easy to switch from a consistency-based semi-supervised learning framework to a contrastive-learining-based semi-supervised learning formulation, by replacing the consistency regularization in \eqref{eq:loss_u} to a contrastive objective.
In this paper, we use the temporal contrastive objective proposed in TCL~\citep{singh2021semi} to further demonstrate the effectiveness of our video data augmentations. 
We refer to \citet{singh2021semi} for more details.
\fi

\subsection{Intra-clip data augmentation}
\label{sec:intra}
\topic{Temporally-coherent photometric and geometric augmentation.}
Moderate photometric (color) and geometric (spatial) variations often do not affect the class semantics (e.g., object classification).
Thus, photometric and geometric augmentation strategies are widely used in supervised~\citep{cubuk2020randaugment}, self-supervised~\citep{chen2020simple,he2020momentum}, and semi-supervised~\citep{sohn2020fixmatch} image classification tasks.
Similar to image classification, videos from the same class also exhibit photometric (color) and geometric (spatial) variations (\figref{motivation}(a-b)).
It is thus natural to apply photometric and geometric data augmentation for video classification.
However, as we validate with an ablation study in \secref{abl}, individually applying state-of-the-art photometric and geometric augmentations (\eg RandAugment~\citep{cubuk2020randaugment}) on each video frame leads to sub-optimal performance.
We conjecture that the random transformations for each frame break the temporal coherency of a video clip.
As a result, the frame-wise inconsistency within video clips may cause adverse effects on the learned representations.
A recent work~\citep{qian2020spatiotemporal} also validates the above assumption in the context of self-supervision video representation learning.
Thus, we apply the same photometric and geometric transformations for every frame to maintain the temporal consistency within a sampled video clip.
Specifically, we sample two basic operations from a pool of photometric and geometric transformations (as in RandAugment~\citep{cubuk2020randaugment}) and then apply them to every frame for each video clip.

\begin{figure}[t]
    \centering
    \includegraphics[width=\linewidth]{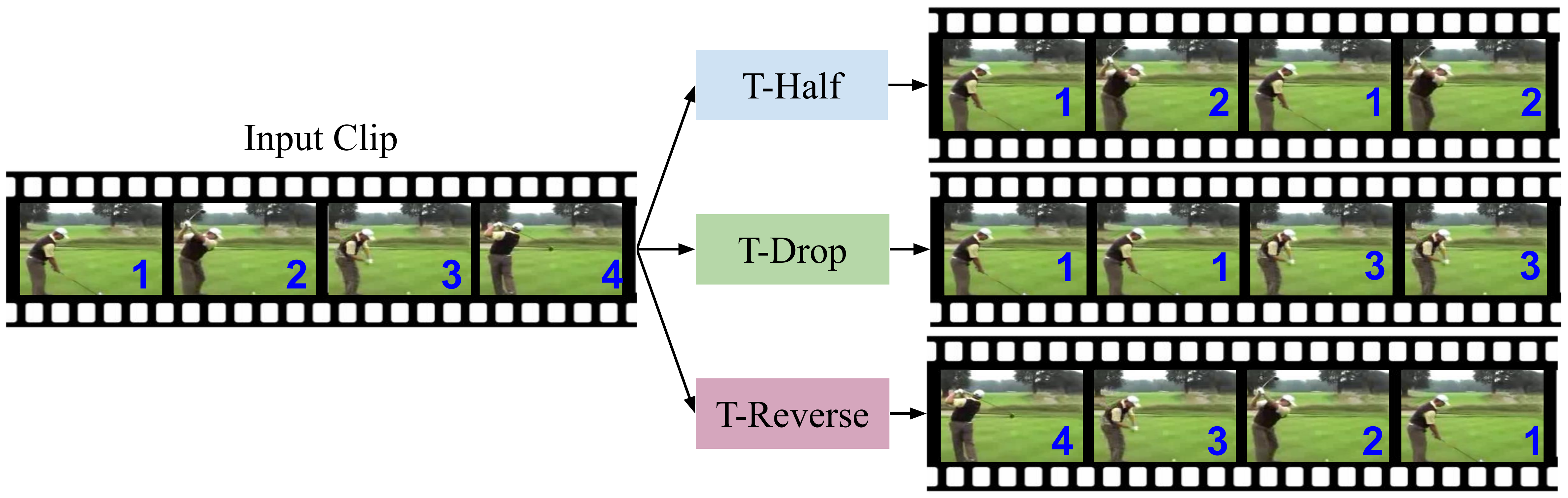}
    \figcaption{Temporal augmentations}{
    We introduce three different temporal transformations: T-Half, T-Drop, and T-Reverse to achieve \emph{temporal occlusion}, \emph{speed}, and \emph{order invariances}.
    The \textbf{\blue{blue numbers}} indicate the frame indices.
    We construct the final temporal augmentation by randomly selecting one of these transformed clips.
    }
    \label{fig:temporal_aug}
\end{figure}

\topic{Temporal augmentation.}
In addition to color space and spatial dimension, videos have a temporal dimension.
The additional temporal dimension significantly increases the variability of video data from many perspectives (\eg speed, sampling rate, temporal order, etc.).
To capture these task-specific representational invariances,
we introduce and study three different types of temporal transformations: (1) T-Half, (2) T-Drop, and (3) T-Reverse.
We illustrate the three temporal augmentations in ~\figref{temporal_aug}.
First, to avoid a video recognition model focusing too much on particular frames instead of understanding the temporal context, we randomly drop some frames within a video clip.
Randomly dropping frames is conceptually similar to the artificial occlusion augmentations, \eg Cutout~\citep{devries2017improved} operations in the spatial dimension in the image domain.
We implement a temporal extension of Cutout in two ways:
1) We randomly discard the second half of a video clip and fill in the empty slots with the first half, T-Half.
2) For each frame in a video clip, we randomly replace it with its previous frame with a probability of $p=0.5$, which we refer to as T-Drop.
In addition to dropping part of the information, T-Drop also simulates \emph{speeding-up} (frame indexes: $[2,3] \rightarrow [1,3]$) and \emph{slowing-down} (frame indexes: $[1,2] \rightarrow [1,1]$ as shown in \figref{temporal_aug}) within a video clip. 
Such temporal augmentation regularizes the video classification model to generate consistent predictions for the two video clips (original and augmented ones) with different speed and temporal occlusion, implicitly encoding the speed/occlusion invariance.
Second, we observe that many actions have cyclic temporal structures and thus can be recognized no matter in the original chronological order or reverse.
To capture this invariance of temporal order,
we transform a video clip by reversing its temporal order, referred to as T-Reverse.
As validated in the ablation study (\secref{abl}), all three operations are beneficial for SSL video action recognition.
Since these operations are complementary to each other, we put them in an operation pool (including the \textit{identity} operation) and randomly sample one for each input video clip.

\topic{Discussions.}
We encourage three types of temporal invariances during model training by applying the aforementioned temporal augmentation.
(1) Invariance to partial temporal occlusion (T-Half and T-Drop):
This invariance enforces models to recognize actions even with partial evidence.
For example, humans can recognize the ``Basketball" action by seeing only the first half (seeing bent arms with a ball, stretching arms but not seeing a ball flying).
(2) Invariance to speed (T-Drop): 
The same action can be performed at a different speed.
A model could also encounter videos containing the same actions but different frame rates.
Hence, encouraging action recognition models to be invariant to different speeds will be helpful.
(3) Invariance to temporal (reverse) order (T-Reverse):
For example, cyclic/order invariant actions such as ``push/pull-ups'', ``punching bags'', and many instrument-playing actions are invariant to reversed order. 
More than 50\% of classes in standard human action recognition benchmarks are cyclic/order-invariant (\eg 57 out of 101 in UCF-101 and 28 out of 51 in HMDB-51).

We validate the effectiveness of these temporal invariances in \tabref{abl}(b). 
Here, we discuss the limitation as well.
First, for fine-grained action recognition~\citep{goyal2017something,mahdisoltani2018effectiveness}, a model may need to utilize the information from all the input frames to make a prediction.
Also, speed difference may play a key role in discriminating the subtle differences between two similar actions (\eg jogging v.s. running).
Partial temporal occlusion invariance and speed invariance could hurt the overall performance in this case.
Second, suppose we want to discriminate action classes with symmetric temporal orders (\eg move an object from left to right v.s. move an object from right to left). 
In that case, encouraging temporal order invariance can be harmful.
However, we find all three temporal invariances helpful for general coarse-grained human action recognition purposes, particularly in a low-label regime.
Note that the users can always decide which temporal invariance to inject on specific video tasks.

\subsection{Cross-clip data augmentation}
\label{sec:cross}
\topic{ActorCutMix.}
There are severe scene representation biases~\citep{li2018resound,Choi-NeurIPS-2019,li2019repair} in the popular action recognition datasets such as UCF-101, HMDB-51, Kinetics-400, Charades~\citep{sigurdsson2016hollywood}, and ActivityNet~\citep{caba2015activitynet}, \etc. 
\ifmarked
\red{Action recognition models trained on these scene-biased datasets tend to leverage the background scene information instead of actual action~\citep{Choi-NeurIPS-2019}.}
\else
Action recognition models trained on these scene-biased datasets tend to leverage the background scene information instead of actual action~\citep{Choi-NeurIPS-2019}.
\fi
These scene-biased action recognition models are likely to fail when tested on the new data with unseen actor-scene combinations.
For example, people can do ``Fencing" either in a gym or in a forest, as shown in the scene invariance block in \figref{motivation}(d).
Training a model on the data consisting of ``Fencing" actions in gyms only will likely fail when tested on the new data consisting of ``Fencing" actions in forests.
To mitigate the scene bias, we propose a new human-centric video data augmentation method, ActorCutMix. 
The operation of the proposed ActorCutMix is similar to CutMix~\citep{yun2019cutmix} in mixing the pixels from different samples to create new training data.
However, our method differs in motivation.
ActorCutMix aims to improve \emph{scene invariance}, regularizing a video classification model to focus on the actor to make predictions (\ie scene debiasing).
In contrast, the goal of CutMix is to achieve \emph{occlusion robustness}, encouraging a model to make correct predictions even when part of the image input is occluded.

\begin{figure}[t]
    \centering
    \includegraphics[width=\linewidth]{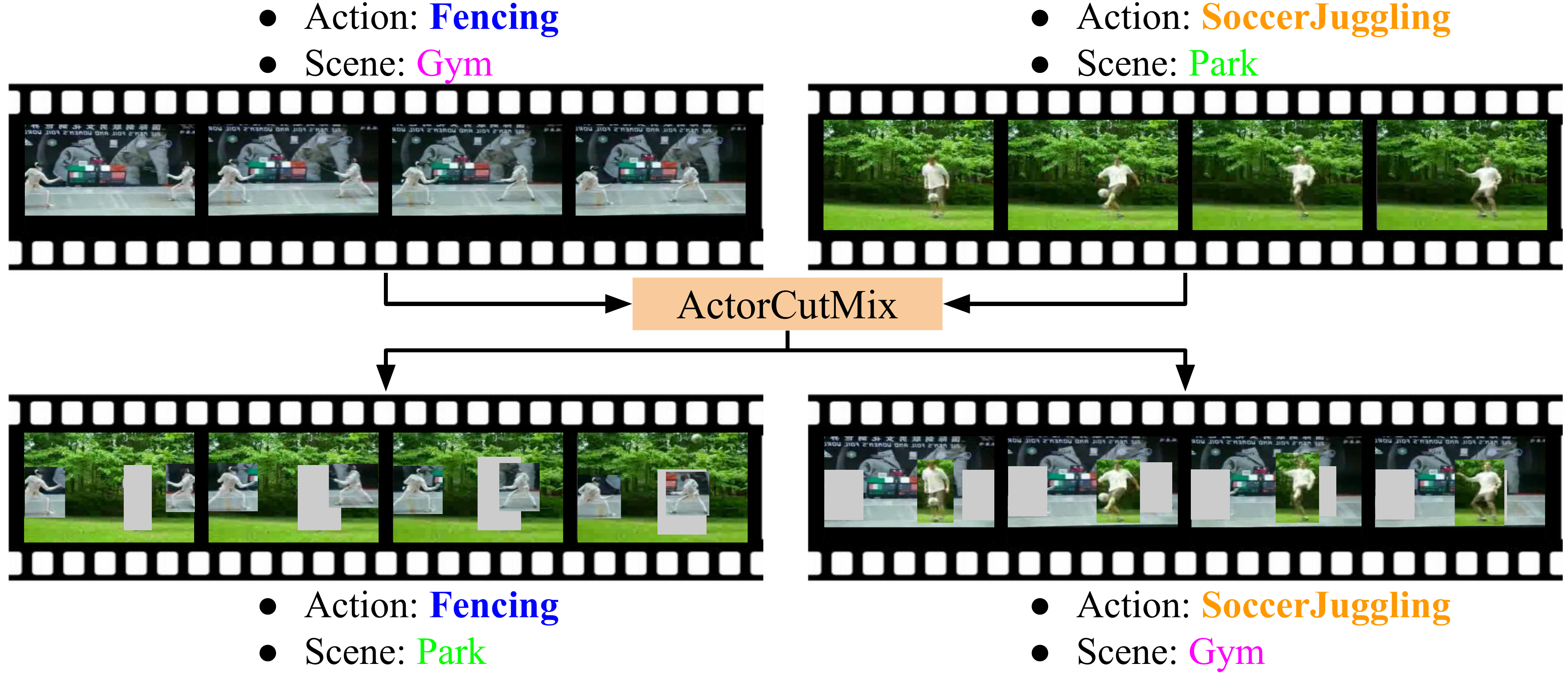}
    \figcaption{ActorCutMix augmentation for scene invariance}{
    We transform two video clips by swapping the background regions of the video clips. 
    We generate human/background masks by running a human detector. 
    }
    \label{fig:actorcutmix_aug}
\end{figure}

As shown in \figref{actorcutmix_aug}, ActorCutMix generates new training examples $(\tilde{x}_A, \tilde{y}_A)$, $(\tilde{x}_B, \tilde{y}_B)$ by swapping the background regions in the two training examples $(x_A, \hat{y}_A)$, and $(x_B, \hat{y}_B)$ in a mini-batch, where $x$ is a video clip and $\hat{y}$ is the corresponding pseudo-label.
We define the swapping operation as follows:

\begin{align}
    \tilde{x}_A &= {m}_A \odot {x}_A + (\textbf{1} - {m}_A) \odot (\textbf{1} - {m}_B)\odot {x}_B, \nonumber \\
    \tilde{y}_A &= \hat{y}_A. %
    \label{eqn:actorcutmix}
\end{align}
Here, $m \in \mathcal{R}^{T\times H\times W\times 3}$ is a binary human mask for a video clip,
with a value of $1$ for human regions and $0$ for the background regions. $\odot$ represents the element-wise multiplication. 
The other augmented training sample and its pseudo label $(\tilde{x}_B, \tilde{y}_B)$ can be generated similarly.
We generate the human mask ${m}$ by running an off-the-shelf human detection algorithm~\citep{he2017mask, yang2019unsupervised}. 
We run the human detector on the video datasets offline and load the cached human bounding boxes during training.
We demonstrate that ActorCutMix significantly improves the semi-supervised action recognition performance in \secref{abl}.

\topic{Discussions.}
ActorCutMix injects the scene invariance into the model, mitigating the reliance on the scene context.
In other words, we encourage a model to recognize the action class by focusing on the actor, regardless of the background scene.
Without proper regularization, deep neural networks tend to learn shortcuts~\citep{devries2017improved,singh2017hide,geirhos2018imagenet,hendricks2018women,yun2019cutmix}. 
Action recognition models could learn spurious correlations between the action and the scene instead of focusing on the human action itself~\citep{Choi-NeurIPS-2019,bahng2020learning},
\ie predicting the action class according to the scene context only.
In this work, we regularize the network to focus on human actions by providing diverse background context for each action.

\topic{Label smoothing.}
In Eqn.~\eqref{eqn:actorcutmix}, we assign the pseudo label of the human action $\hat{y}_A$ to the video clip $A$ by assuming a perfect human detector.
In other words, we aim to recognize the \emph{human action} instead of the background scene.
However, human detectors are imperfect in general.
Due to missing/false-positive human detections, augmented video clips could potentially contain humans performing different actions from different clips.
Therefore, to prevent model confusion, we apply label smoothing with a higher weight for the label of the (potentially corrupted) human action $\hat{y}_A$ and a lower weight for the label of the (potentially corrupted) background scene $\hat{y}_B$ as follows. 

\begin{align}
    \tilde{y}_A &= \lambda \hat{y}_A + (1-\lambda) \hat{y}_B. \\ %
    \lambda &= 1-\left|1- \frac{\sum{m_A}}{THW} \right|^{\alpha}.
    \label{eqn:labelsmooth}
\end{align}
Here, ${\sum{m_A}}/{THW \in [0,1]}$ is the foreground (human) ratio, and $\alpha$ is a hyperparameter that controls the influence of the pseudo label of the human region. We empirically find $\alpha=4$ yields good results.

We emphasize that ActorCutMix and CutMix have different purposes of label smoothing.
Label smoothing in CutMix aims to provide multiple labels for a single training image. 
For example, a training image consists of information from two different classes (\eg a dog and a cat).
In contrast, we smooth labels in ActorCutMix to mitigate data corruption due to the missing/false-positive human detections.
The proposed label smoothing can further boost performance as shown in \tabref{abl}(c).

\topic{Limitations.}
Scene invariance also has its limitation.
For example, if our goal is to discriminate different actions in the same background context (\eg the Diving48 dataset~\citep{li2018resound}), ActorCutMix makes no difference.
Also, the current combination of the actors and the scenes could cause visual artifacts that may be harmful, as shown in other copy and paste-based augmentations~\citep{devries2017improved,yun2019cutmix,kim2020puzzle,Yoo-2020-CVPR}.
More accurate segmentation of the actors and objects may further improve the performance.
We leave it as a future direction.

\subsection{Combining different data augmentations}
\label{sec:combine}

We provide the algorithm outline of our strong data augmentation strategy for video in \algref{alg:aug}.

\topic{Combining photometric-geometric and temporal augmentations.}
We combine the photometric-geometric and temporal augmentations by cascading both of them.
The cascaded combination can be regarded as a spatial-temporal counterpart of RandAugment~\citep{cubuk2020randaugment}: Randomly sample two operations from photometric and geometric augmentation operations, and then sample one operation from the temporal augmentation operations.
We refer this combined augmentation as \emph{intra-clip augmentation}.

\topic{Combining intra-clip and cross-clip augmentations.}
To combine intra-clip augmentation (photometric-geometric-temporal) and cross-clip augmentation (ActorCutMix), we propose randomly applying either one of them for each mini-batch.
We validate the effectiveness of this combination in \secref{abl}.

\begin{algorithm}[ht]
    \SetKwInOut{Input}{Input}
    \SetKwInOut{Return}{Return}
    
    \Input{A mini-batch of unlabeled video clips $X$ and the corresponding human mask $M$}
    Draw a sample $p$ from uniform distribution $U(0, 1)$\\
    \textbf{If} $p$ $>$ 0.5 \\
    \quad Reverse the order of the batch dimension to get $X'$ and $M'$ \\
    \quad $\tilde{X}, \tilde{X'} = \textbf{ActorCutMix}(X, X', M, M')$ \\

    \textbf{else} \\
    \quad \textbf{for} each video clip $x$ in $X$\\
    \quad \quad Sample $op_1$, $op_2$ from photometric-geometric op pool, $op_3$ from temporal op pool \\
    \quad \quad $\tilde{x} = \textbf{PhotometricGeometricAug}(x, op_1, op_2)$ \\
    \quad \quad $\tilde{x} = \textbf{TemporalAug}(\tilde{x}, op_3)$ \\
    \quad \textbf{end for} \\
    \textbf{end if}\\
    \Return{Strongly-augmented video clips $\tilde{X}$}
    \caption{Strong video data augmentation}
    \label{alg:aug}
\end{algorithm}
\section{Experimental Results}
\label{sec:results}

To validate our strong data augmentation strategy, we conduct three types of experiments: i) consistency-based semi-supervised learning in \secref{consistency_exp}, ii) contrastive learning-based semi-supervised learning in \secref{comp_TCL}, iii) fully-supervised learning in \secref{full_sup}.

\begin{table}[ht]
\centering

\caption{\tb{Hyper-parameters.}
}

\resizebox{\linewidth}{!}{

\begin{tabular}{clc}
\toprule
Symbol & Description & Value  \\

\midrule

$\tau$ & Pseudo label threshold (Eq.~(\red{2})) & 0.95 \\

$\lambda_u$ & Unlabeled loss weight (Eq.~(\red{4})) & 1.0 \\

$\alpha$ & Scaling factor for label smoothing (Eq.~(\red{7})) & 4.0 \\

\bottomrule
\end{tabular}

}

\label{tab:hyper}
\end{table}

\begin{table*}[t]
\centering
\caption{\tb{Ablation study.}
We show results from an R(2+1)D ResNet-34 model on the 20\% label split of the UCF-101 dataset.
}

\mpage{0.34}{(a) Temporally-coherent photometric-geometric augmentation}\hfill
\mpage{0.30}{(b) Temporal augmentation}\hfill
\mpage{0.32}{(c) ActorCutMix augmentation}\hfill
\\
\mpage{0.32}{
\begin{tabular}{lcc}
\toprule
Strategy & Top-1 acc. \\
\midrule
Supervised baseline & 38.91 \\
\midrule
Per-frame & 44.17 \\
Temporally-coherent & \textbf{53.37} \\
\bottomrule
\end{tabular}
}
\mpage{0.32}{
\resizebox{0.75\linewidth}{!}{
\begin{tabular}{lcc}
\toprule
Strategy & Top-1 acc. \\
\midrule
Sup. baseline & 38.91 \\
Speed up and down & 41.66\\
\midrule
T-Half & 42.77 \\
T-Drop & 43.14 \\
T-Reverse & 43.40 \\
\midrule
TemporalAll & \textbf{44.07} \\
\bottomrule
\end{tabular}
}
}
\mpage{0.32}{
\resizebox{1.05\linewidth}{!}{
\begin{tabular}{lcc}
\toprule
Strategy & Detection & Top-1 acc. \\
\midrule
Sup. baseline & - & 38.91 \\
Bkg. CutMix & \cite{he2017mask} & 41.87 \\
CutMix & - & 43.27 \\
\midrule
w/o smoothing & \cite{he2017mask} & 42.82 \\
w/ smoothing & \cite{he2017mask} & \textbf{45.28} \\
\midrule
w/ unsup. det. & \cite{yang2019unsupervised} & 45.12 \\
\bottomrule
\end{tabular}
}
}
\\
\vspace{2mm}

\mpage{0.30}{(d) Combining intra-clip augs.}
\mpage{0.38}{(e) Combining intra- and cross-clip augs.}\hfill
\mpage{0.29}{(f) Different SSL frameworks}\hfill
\\
\mpage{0.32}{
\begin{tabular}{lcc}
\toprule
Strategy & Top-1 acc. \\
\midrule
Supervised baseline & 38.91 \\
\midrule
Randomly sample one & 53.37 \\
Cascaded & \textbf{54.48} \\
\bottomrule
\end{tabular}
}
\hfill
\mpage{0.32}{
\begin{tabular}{lcc}
\toprule
Strategy & Top-1 acc. \\
\midrule
Supervised baseline & 38.91 \\
Intra only & 54.48 \\
\midrule
Cascaded & 50.89 \\
Randomly sample one & \textbf{56.73} \\
\bottomrule
\end{tabular}
}
\hfill
\mpage{0.32}{
\begin{tabular}{lcc}
\toprule
Method & FixMatch & UDA \\
\midrule
Vanilla & 44.17 & 43.11 \\
Vanilla+temp.-co. & 53.37 & 50.20 \\
Ours & \textbf{56.73} & \textbf{54.80} \\
\bottomrule
\end{tabular}
}
\label{tab:abl}
\end{table*}

\subsection{Experimental results on consistency-based SSL}
\label{sec:consistency_exp}
We validate our strong data augmentation strategy by plugging it into consistency-based semi-supervised learning frameworks (\ie FixMatch~\citep{sohn2020fixmatch} and UDA~\citep{xie2019unsupervised}). 
We start with describing experimental setting in \secref{setup}, then we show ablation experiments in \secref{abl}. 
Comparison with state of the arts (\secref{comp_sota}) and error analysis (\secref{error_analysis}) are followed. 

\subsubsection{Experimental setup}
\label{sec:setup}
\topic{Dataset.}
We conduct experiments on the public action recognition benchmarks: UCF-101~\citep{soomro2012ucf101}, HMDB-51~\citep{kuehne2011hmdb}, and Kinetics-100~\citep{jing2020videossl,kay2017kinetics}.
UCF-101 consists of $13,320$ videos with 101 action classes.
HMDB-51 consists of $6,766$ videos with 51 action classes.
The full Kinetics dataset consists of $300K$ videos with 400 classes.
We use a subset, Kinetics-100, consisting of $90K$ videos with 100 classes. 
We split the datasets following \citet{jing2020videossl} to conduct semi-supervised training based on the state-of-the-art FixMatch framework~\citep{sohn2020fixmatch}.

\topic{Evaluation metrics.}
For all the datasets, we report top-1 accuracy for quantitative comparison.

\topic{Compared methods.}
As a first baseline, we train a model with only the labeled data using standard/weak video data augmentations (\eg random scaling, horizontal flipping, etc.).
We call it as \emph{supervised baseline}.
In the low-label setting,
we also compare our method with VideoSSL~\citep{jing2020videossl} and several semi-supervised methods adapted to video (\ie PseudoLabel~\citep{lee2013pseudo}, MeanTeacher~\citep{tarvainen2017mean}, S4L~\citep{zhai2019s4l}).

\topic{Implementation details.}
We implement our method on top of the publicly available mmaction2 codebase~\citep{2020mmaction2}.
Unless specified, we use the R(2+1)D model~\citep{Tran-CVPR-2018} with ResNet-34~\citep{he2016deep} as the feature extraction backbone.
To better understand the effect of our augmentation techniques, we initialize the model with \emph{random weights} (as opposed to using models with supervised pre-training on ImageNet and/or Kinetics).
We sample eight frames from each video randomly and uniformly to construct a clip with an eight-frame sampling interval.
We use a batch size of 16 clips for the supervised learning baseline for each GPU.
We use a mini-batch of five clips from labeled data and five clips from unlabeled data for each GPU for our semi-supervised learning method.
We train our models using 8 RTX 2080 Ti GPUs.

We use SGD with momentum as our optimizer, with an initial learning rate of $0.2$, a momentum value of $0.9$, and a weight decay value of $1e-4$.
We use the cosine annealing policy for learning rate decay.
We also adopt synchronous batch normalization across eight GPUs.
For the UCF-101 dataset~\citep{soomro2012ucf101} and Kinetics-100 dataset~\citep{kay2017kinetics,jing2020videossl}, we train our method for 360 epochs with respect to unlabeled data.
For the HMDB-51 dataset~\citep{kuehne2011hmdb}, we train our method for 600 epochs with respect to unlabeled data.
We list the values of the other hyper-parameters as in \tabref{hyper}.

\subsubsection{Ablation study}
\label{sec:abl}
In the following, we validate each design choice of our strong data augmentation strategy. 
We conduct ablation experiments on the 20\% label split of the UCF-101 dataset~\citep{soomro2012ucf101} using a R(2+1)D ResNet-34 model in the semi-supervised learning setting based on FixMatch~\citep{sohn2020fixmatch}.

\topic{Temporally-coherent photometric-geometric augmentation.}
We first conduct an ablation experiment to study the necessity of applying \emph{temporally-coherent} photometric-geometric augmentation. 
As shown in \tabref{abl}(a), although applying photometric-geometric augmentation individually in a frame-by-frame manner improves upon the supervised baseline ($38.91\% \rightarrow 44.17\%$), using temporally-coherent augmentation further enhances the performance by a large margin ($44.17\% \rightarrow 53.37\%$).
The results validate our hypothesis that preserving frame-wise consistency in the augmented videos leads to improved results.

\topic{Temporal augmentation.}
Next, we study the effectiveness of temporal augmentation and its atomic operations.
\ifmarked
\red{As shown in \tabref{abl}(b), all of the atomic operations, \ie T-Half, T-Drop, T-Reverse, are beneficial to the recognition accuracy.}
\else
As shown in \tabref{abl}(b), all of the atomic operations, \ie T-Half, T-Drop, T-Reverse, are beneficial to the recognition accuracy.
\fi
The results validate our motivation: Temporal occlusion, speed, and order invariances are beneficial for coarse-grained human action recognition, improving the data efficiency.
With all the atomic temporal operations combined, the final temporal augmentation further improves the accuracy.
\ifmarked
\red{In addition, we compare our temporal augmentation (TemporalAll) with an existing speed-up and down augmentation baseline~\citep{singh2021semi}. 
Our temporal augmentation shows favorable performance compared to the speed-up and speed-down augmentation ($44.07\%$ vs.$41.66\%$).}
\else
In addition, we compare our temporal augmentation (TemporalAll) with an existing speed-up and down augmentation baseline~\citep{singh2021semi}. 
Our temporal augmentation shows favorable performance compared to the speed-up and speed-down augmentation ($44.07\%$ vs.$41.66\%$).
\fi

\topic{ActorCutMix augmentation.}
In \tabref{abl}(c), ActorCutMix without label smoothing shows moderate improvement over the 
\ifmarked
\red{supervised}
\else
supervised 
\fi
baseline.
With label smoothing, ActorCutMix shows even more significant performance improvement. 
The results imply that label smoothing can mitigate data corruption caused by the missing/false-positive human detections.
\ifmarked
\red{When compared with CutMix~\citep{yun2019cutmix} ($43.27\%$) and Background CutMix ($41.85\%$) where we replace the background instead of replacing humans, ActorCutMix ($45.26\%$) shows significant accuracy improvement.
The results validate that capturing scene invariance improves action recognition accuracy.}
\else
When compared with CutMix~\citep{yun2019cutmix} ($43.27\%$) and Background CutMix ($41.85\%$) where we replace the background instead of replacing humans, ActorCutMix ($45.26\%$) shows significant accuracy improvement.
The results validate that capturing scene invariance improves action recognition accuracy.
\fi
To study the effect of supervision for human detection, we replace the supervised detector~\citep{he2017mask} with an unsupervised detector~\citep{yang2019unsupervised} trained on the UCF-101 dataset. 
ActorCutMix with an unsupervised detector slightly underperforms ($45.29\% \rightarrow 45.12\%$) than the ActorCutMix with a supervised detector (both are with label smoothing).
We use the supervised human detector for the rest of the experiments.

\topic{Combining photometric-geometric and temporal augmentations.}
In this experiment, we study how to combine photometric-geometric augmentation and temporal augmentation.
We compare the cascaded strategy with an alternative combination strategy: sample only one of them and apply it for a video clip.
As shown in \tabref{abl}(d), the cascaded strategy leads to better performance.

\topic{Combining intra-clip and cross-clip augmentations.}
We study how to combine both intra-clip (photometric-geometric-temporal) and cross-clip (ActorCutMix) data augmentations.
A straightforward approach is cascading these two types of augmentations. 
\ifmarked
\red{As shown in \tabref{abl}(e), we find that the cascading intra- and cross-clip augmentations is even \emph{worse} than applying the intra-clip augmentation without ActorCutMix ($50.89\%$ vs. $54.48\%$).}
\else
As shown in \tabref{abl}(e), we find that the cascading intra- and cross-clip augmentations is even \emph{worse} than applying the intra-clip augmentation without ActorCutMix ($50.89\%$ vs. $54.48\%$).
\fi
Our intuition is that cascading intra-clip and cross-clip augmentation produces severely distorted clips that no longer resemble natural videos. 
Randomly applying only one data augmentation selected from intra-clip \emph{or} cross-clip for each input video clip gives the best top-1 accuracy ($56.73\%$). Hence, we use random sampling for the rest of the paper.

\topic{Improvement over vanilla semi-supervised frameworks.}
Lastly, we show that our strong data augmentation is method-agnostic. We plug our strong augmentation into the unlabeled branches of two state-of-the-art consistency-based semi-supervised learning frameworks, FixMatch~\citep{sohn2020fixmatch} and UDA~\citep{xie2019unsupervised}.
In \tabref{abl}(f), the two semi-supervised learning frameworks with per-frame augmentation are denoted as vanilla. The vanilla+temp.-co. denotes we use temporally coherent photometric/geometric augmentations for strong augmentation. Our strong augmentation consistently improves the performance by a large margin. We show improvement over another semi-supervised action recognition framework, TCL \citep{singh2021semi} in \secref{comp_TCL}.

\topic{Strong augmentation on the labeled branch.}
Currently, the \emph{strong} data augmentation is only applied to the unlabeled branch by default, as shown in \figref{overview}.
Here, we also conduct an experiment applying the strong augmentation on the labeled branch.
It achieves a similar performance (56.46\%) as our default setting (56.73\%).
We conjecture that the strong augmentations have already injected the visual invariances in the unlabeled branch. 
Additional strong augmentation for the labeled branch may not add extra visual invariances to the model.
Therefore, the performance difference is negligible.
For better training efficiency, we choose to use the default setting.

\topic{Different initialization.}
We investigate whether our strong data augmentation strategy
can still improve over the \emph{supervised baseline}, given strong pre-trained model weights as an initialization.
As shown in \tabref{abl_init}, while the improvement over the supervised baseline is not as significant as the boost in the train-from-scratch setting, the proposed method can still achieve a sizable improvement (72.40\% $\rightarrow$ 77.37\%).
The results validate the general applicability of the strong data augmentation in practical scenarios.
\begin{table}[ht]
\centering

\caption{\tb{Effect of using different initialization.}
}

\begin{tabular}{lccc}
\toprule
& \multicolumn{2}{c}{Initialization} \\
\cmidrule{2-3}
Method & Random & Kinetics-400  \\

\midrule

Supervised & 38.91  & 72.40 \\ 

Ours & \textbf{56.73} & \textbf{77.37} \\

\bottomrule
\end{tabular}

\label{tab:abl_init}
\end{table}

\subsubsection{Comparing with the state of the arts}
\label{sec:comp_sota}
\begin{table*}[htbp]
\centering
\caption{\tb{Comparison with the state of the arts.}
All the methods use a 3D ResNet-18 model.
The best performance is in \best{bold} and the second best is \second{underlined}.
}

\resizebox{\linewidth}{!}{
\begin{tabular}{lccccccccccc}
\toprule
& w/ ImageNet & Kinetics-100 && \multicolumn{4}{c}{UCF-101} && \multicolumn{3}{c}{HMDB-51}\\
\cmidrule{3-3}
\cmidrule{5-8}
\cmidrule{10-12}
Method & distillation & 20\% && 50\% & 20\% & 10\% & 5\% && 60\% & 50\% & 40\% \\
\midrule
PseudoLabel~\citep{lee2013pseudo} & - 
& 48.0
&& 47.5 & 37.0 & 24.7 & 17.6
&& 33.5 & 32.4 & 27.3
\\
MeanTeacher~\citep{tarvainen2017mean} & - 
& 47.1 
&& 45.8 & 36.3 & 25.6 & 17.5
&& 32.2 & 30.4 & 27.2
\\
S4L~\citep{zhai2019s4l} & - 
& 51.1 
&& 47.9 & 37.7 & 29.1 & 22.7 
&& 35.6 & 31.0 & 29.8
\\
VideoSSL~\citep{jing2020videossl} & \checkmark 
& {57.7} 
&& {54.3} & {48.7} & \second{42.0} & \second{32.4}
&& {37.0} & {36.2} & {32.7}
\\
Ours & - 
& \second{61.2} 
&& \second{59.9} & \second{51.7} & {40.2} & {27.0}
&& \second{38.9} & \second{38.2} & \second{32.9}
\\
Ours & \checkmark 
& \best{68.7} 
&& \best{64.7} & \best{57.4} & \best{53.0} & \best{45.1}
&& \best{40.8} & \best{39.5} & \best{35.7}
\\
\bottomrule
\end{tabular}
}

\label{tab:main_all}
\end{table*}

Next, we compare our method on several established benchmarks, following the same data splits as in \citet{jing2020videossl}.
As shown in \tabref{main_all}, our method consistently achieves favorable performance compared with other approaches with the same amount of supervision.
On the UCF-101 dataset, VideoSSL~\citep{jing2020videossl} achieves better performance in the extremely low label ratios (10\% and 5\%) than ours while being worse in other label ratios.
We hypothesize that the ImageNet pre-trained model's knowledge distillation plays a crucial role in the extremely low-label regime. 
The improvement from the distillation might quickly become marginal as the label ratio increases.

\topic{ImageNet Knowledge Distillation.}
Following VideoSSL~\citep{jing2020videossl}, we adopt an ImageNet pre-trained ResNet-18 to compute a $1,000$ dimensional ImageNet class probability vector for each frame of all the training videos offline.
Accordingly, we add a classification head to predict the ImageNet probability of each video clip on top of the feature extraction backbone.
We randomly select one frame from each (weakly-augmented) video clip during training time and then use its corresponding ImageNet probability as a soft pseudo label for a cross-entropy loss.
The soft pseudo label provides an additional supervisory signal to the feature backbone.
As shown in \tabref{main_all}, our method's performance significantly improves with ImageNet distillation compared to our method without ImageNet distillation ($26.1\%\rightarrow49.5\%$ in $5\%$ label split and $39.9\%\rightarrow52.7\%$ in $10\%$ label split of the UCF-101 dataset).
Our method with ImageNet distillation shows a favorable performance when compared to VideoSSL~\citep{jing2020videossl}.

\topic{Comparison with self-supervised learning approaches.}
With the same amount of data, one can pre-train the model using all the available data (self-supervised learning) and then fine-tune the model with a small amount of labeled data instead of conducting semi-supervised learning.
Thus, we establish another type of baselines by fine-tuning the self-supervised video pre-trained models on the labeled data.
We choose three recent state-of-the-art self-supervised methods for video representation, \ie VCOP~\citep{xu2019self}, DPC~\citep{han2019video}, and MemDPC~\citep{han2020memory}.
As shown in \tabref{main_selfsup}, our method achieves promising results when compared with these self-supervised approaches across three datasets.
Note that these self-supervised learning approaches are complementary to existing semi-supervised learning frameworks. For example, one can first pre-train the feature backbone with a self-supervised learning objective and then fine-tune the network in a semi-supervised manner.

\begin{table*}[htbp]
\centering
\caption{\tb{Comparison with self-supervised learning approaches}.
The best performance is in \best{bold} and the second best is \second{underlined}.
}

\resizebox{\linewidth}{!}{
\begin{tabular}{lccccccccccc}
\toprule
&& Kinetics-100 && \multicolumn{4}{c}{UCF-101} && \multicolumn{3}{c}{HMDB-51}\\
\cmidrule{3-3}
\cmidrule{5-8}
\cmidrule{10-12}
Method & Network & 20\% && 50\% & 20\% & 10\% & 5\% && 60\% & 50\% & 40\% \\
\midrule
VCOP~\citep{xu2019self} & R(2+1)D ResNet-10  
& 50.9
&& \second{67.9} & \second{53.1} & 31.1 & 14.4
&& 27.8 & 26.5 & 25.7
\\
DPC~\citep{han2019video} & R(2+3)D ResNet-18 
& 59.2
&& 53.1 & 38.9 & 25.3 & 19.2
&& 37.2 & 33.2 & 31.6
\\
MemDPC~\citep{han2020memory} & R(2+3)D ResNet-18 
& 48.5
&& 60.5 & 47.8 & 31.2 & \best{27.3}
&& \second{40.6} & 32.8 & 29.7
\\
\midrule
Ours & 3D ResNet-18 
& \second{61.2} 
&& {59.9} & {51.7} & \best{40.2} & \second{27.0}
&& {38.9} & \second{38.2} & \second{32.9}
\\
Ours & R(2+1)D ResNet-34
&  \best{65.8}
&& \best{68.7} & \best{56.7} & \second{39.7} & {22.8}
&& \best{41.7} & \best{39.2} & \best{34.6}
\\
\bottomrule
\end{tabular}
}

\label{tab:main_selfsup}
\end{table*}

\subsubsection{Improvement over Supervised Baseline and Error Analysis}
\label{sec:error_analysis}
\begin{figure}[htbp]
\mpage{0.48}{
\includegraphics[width=\linewidth]{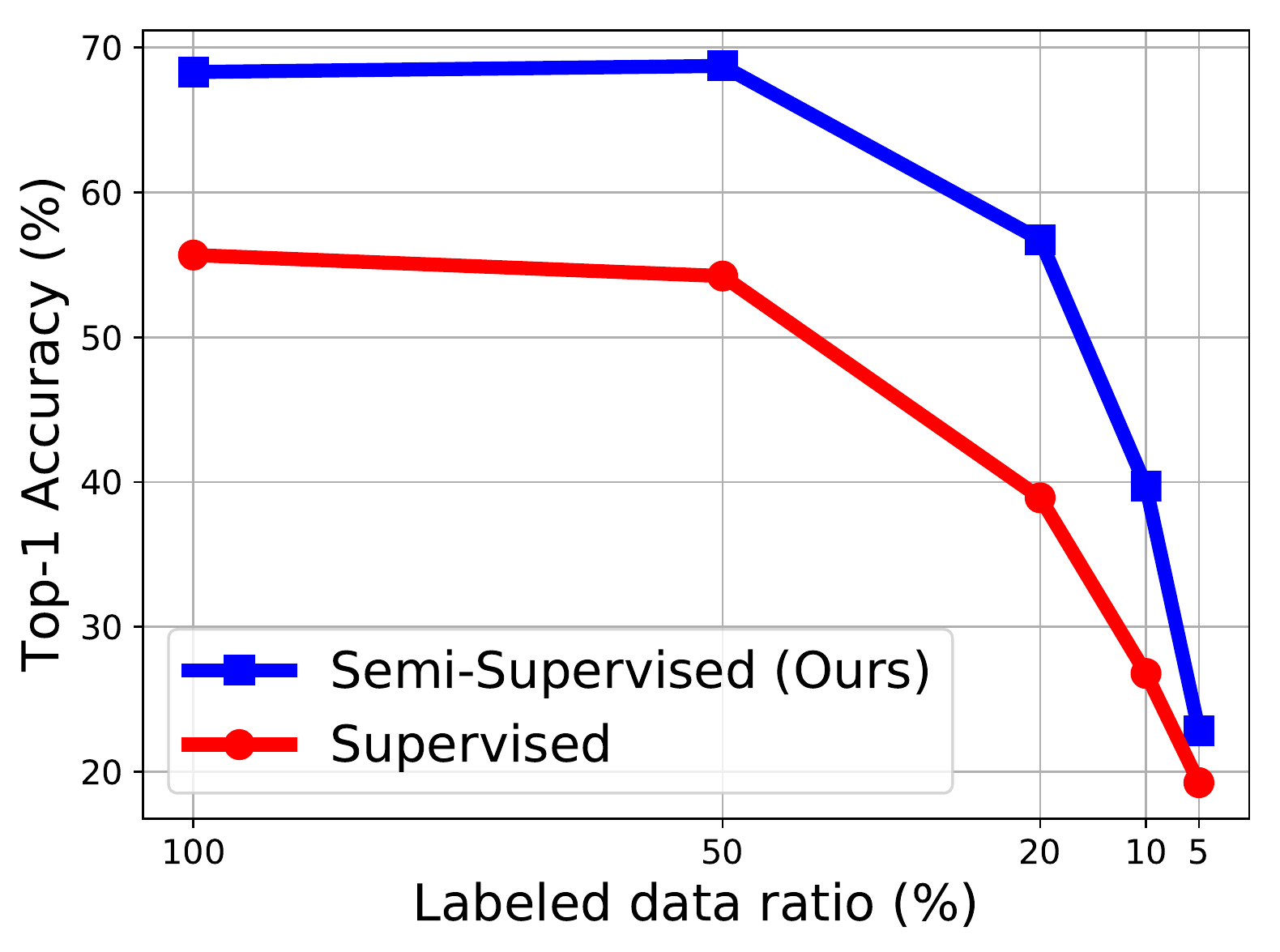}
}\hfill
\mpage{0.48}{
\includegraphics[width=\linewidth]{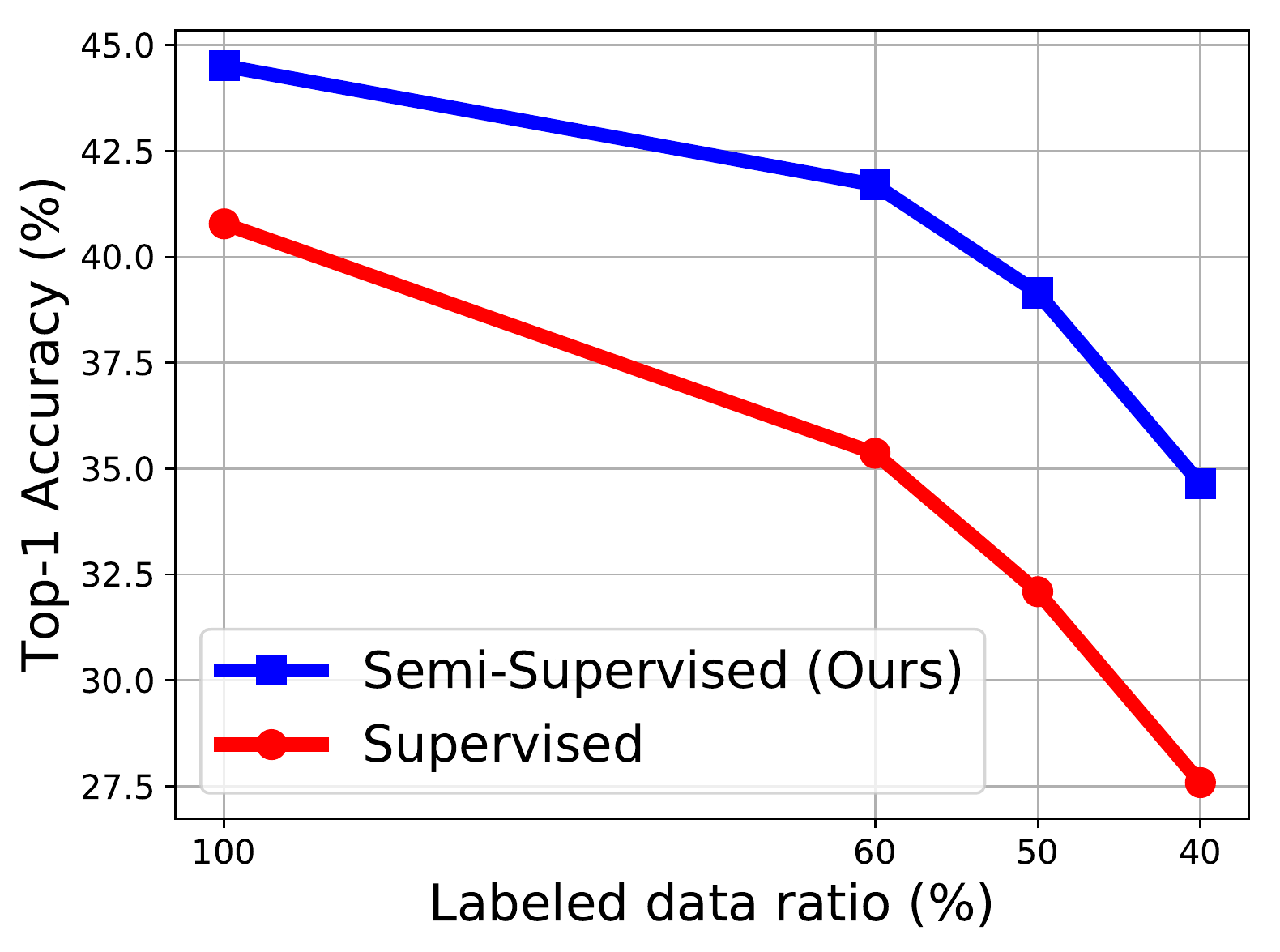}
}\hfill
\mpage{0.48}{(a) UCF-101}\hfill
\mpage{0.48}{(b) HMDB-51}\hfill
\\
\figcaption{Improvement over the supervised baseline}{
We train R(2+1)D ResNet-34 models from scratch.
}
\label{fig:improve_baseline}
\end{figure}

In this section, we plug our strong data augmentation into FixMatch~\citep{sohn2020fixmatch} and compare it with the supervised baseline, where only labeled video data is used during training (labeled branch only in \figref{overview}). 
Note that we train both our model and the supervised baseline from scratch. 
As shown in \figref{improve_baseline}, our method consistently outperforms the supervised baseline by a large margin across different label ratios in both the UCF-101 and HMDB-51 datasets.
The performance improvement highlights the efficacy of injecting video representational invariances with our strong data augmentation strategy.

\begin{figure*}[t]
    \centering
    \includegraphics[width=\linewidth]{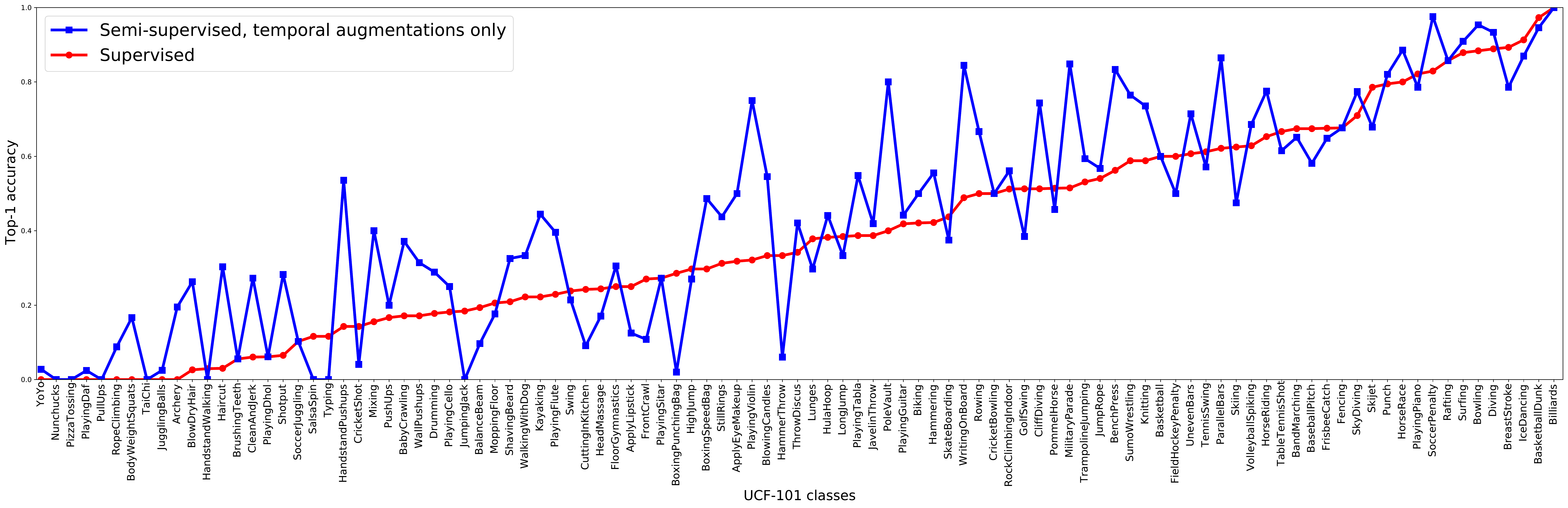}
    \mpage{0.98}{(a) Temporal augmentations only, semi-supervised vs. Supervised baseline}
    \includegraphics[width=\linewidth]{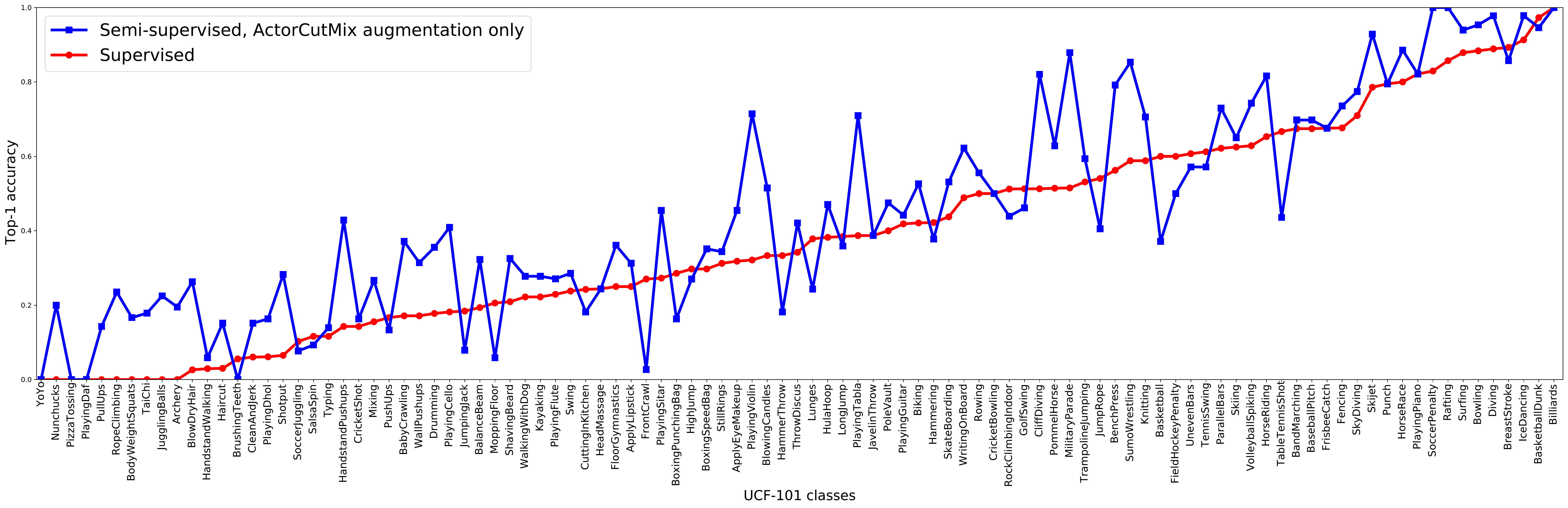}
    \mpage{0.98}{(b) ActorCutMix augmentation only, semi-supervised vs. Supervised baseline}
    \includegraphics[width=\linewidth]{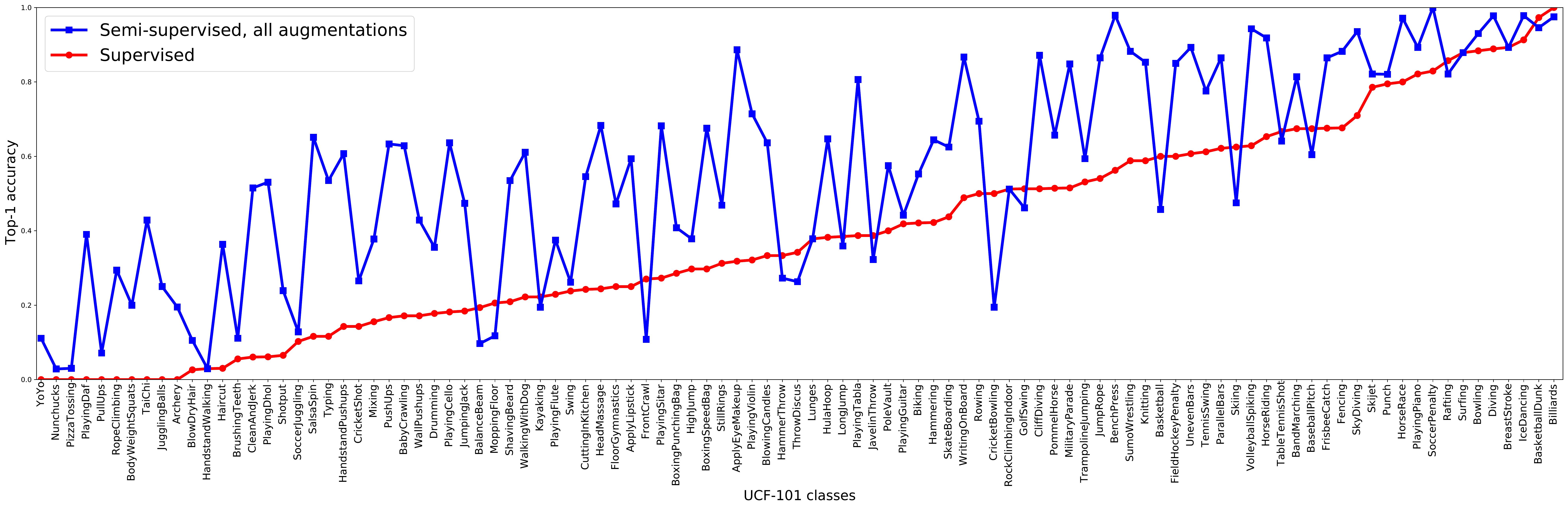}
    \mpage{0.98}{(c) All augmentations, semi-supervised vs. Supervised baseline}

    \figcaption{Class accuracy}{We compare per-class accuracies of the supervised baseline and the semi-supervised model with (a) temporal augmentations only, (b) ActorCutMix augmentation only, and (c) our final all augmentations on the $20\%$ label split of the UCF-101. 
    Classes are sorted in ascending order of the supervised baseline accuracy. Best viewed with zoom and color.
    }
    \label{fig:cls_acc}
\end{figure*}

\begin{figure*}[t]
\mpage{0.32}{
\includegraphics[width=\linewidth]{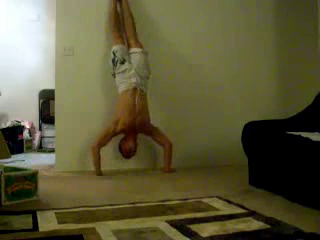}
}\hfill
\mpage{0.32}{
\includegraphics[width=\linewidth]{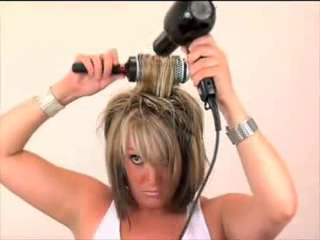}
}\hfill
\mpage{0.32}{
\includegraphics[width=\linewidth]{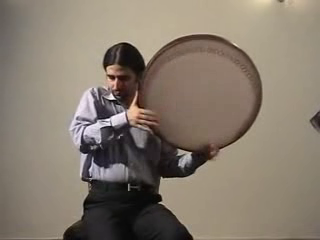}
}\hfill
\mpage{0.32}{
(a) \\
GT: HandstandPushups\\
Baseline pred.:\red{RockClimbIndoor}\\
Temp. aug only pred.:\green{HandstandPushups}\\
}\hfill
\mpage{0.32}{
(b) \\
GT: BlowDryHair\\
Baseline pred.: \red{ApplyLipstick}\\
ActorCutMix only pred.: \green{BlowDryHair}\\
}\hfill
\mpage{0.32}{
(c) \\
GT: PlayingDaf\\
Baseline pred.:\red{PlayingFlute}\\
All aug. pred.:\green{PlayingDaf}\\
}\hfill
\vspace{1em}
\mpage{0.32}{
\includegraphics[width=\linewidth]{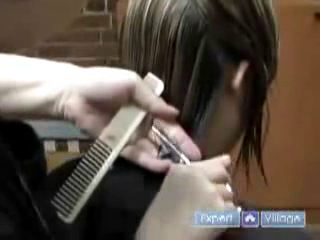}
}\hfill
\mpage{0.32}{
\includegraphics[width=\linewidth]{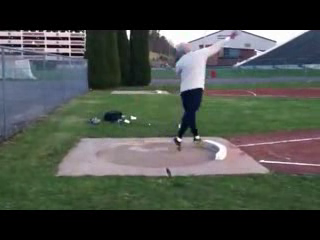}
}\hfill
\mpage{0.32}{
\includegraphics[width=\linewidth]{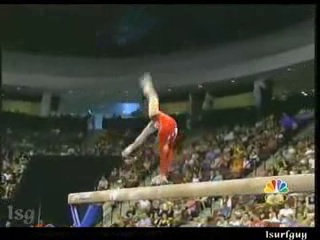}
}\hfill
\mpage{0.32}{
(d) \\
GT: HairCut\\
Baseline pred.:\red{BlowDryHair}\\
Temp. aug only pred.:\green{HairCut}\\
}\hfill
\mpage{0.32}
{
(e) \\
GT: ShotPut\\
Baseline pred.: \red{JavelinThrow}\\
ActorCutMix only pred.: \green{ShotPut}\\
}\hfill
\mpage{0.32}{
(f) \\
GT: BalanceBeam\\
Baseline pred.:\red{UnevenBars}\\
All aug. pred.:\green{BalanceBeam}\\
}\hfill
\figcaption{Visual examples showing the improvement over the supervised baseline}{
UCF-101 examples misclassified by the baseline but are correctly classified with the proposed augmentations. Correct/incorrect predictions are shown in \green{green}/\red{red}.
}
\label{fig:vis}

\end{figure*}

In \figref{cls_acc}, we visualize the class accuracy curves of the supervised baseline (\red{red} curve) and our semi-supervised model (\blue{blue} curves) with our (i) temporal augmentations only, 
(ii) ActorCutMix augmentation only, and 
(iii) final all augmentations, on 20\% label split of the UCF-101 dataset. 
We sort the classes in ascending order of the baseline accuracy. 
\ifmarked
\red{We show visual examples to demonstrate the improvement of our proposed augmentations compared to the supervised baseline in \figref{vis}.}
\else
We show visual examples to demonstrate the improvement of our proposed augmentations compared to the supervised baseline in \figref{vis}.
\fi

\ifmarked
\red{Temporal augmentations can reduce the error rate for the classes such as ``HandstandPushups" and ``Haircut". 
The baseline model confused ``HandstandPushups" with ``RockClimbIndoor". }
\else
Temporal augmentations can reduce the error rate for the classes such as ``HandstandPushups" and ``Haircut". The baseline model confused ``HandstandPushups" with ``RockClimbIndoor". 
\fi
However, our temporal augmentation significantly reduces this type of error ($28.6\% \rightarrow 0\%$).
\ifmarked
\red{Please see \figref{vis}(a),}
\else
Please see \figref{vis}(a),
\fi
temporal augmentations such as T-Half and T-Drop provide a model with less information. 
Therefore, they could encourage the model to focus more on the fine-grained details related to human actions, such as human pose. 
Compared to the baseline, ActorCutMix reduces the error rates of classes such as ``BlowDryHair" and ``ShotPut." 
The baseline is confused the classes ``BlowDryHair" with ``ApplyLipstick," which has similar scenes, \eg, bedroom and restroom. 
In contrast, ActorCutMix significantly reduces this type of error, ($18.4\% \rightarrow 7.9\%$) 
\ifmarked
\red{as shown in \figref{vis} (b).}
\else
as shown in \figref{vis} (b).
\fi
The results imply that ActorCutMix provides various scenes and results in the scene invariance of our model. 
Additional invariances to photometric/geometric transformations help our model not to use shortcuts.
Using all augmentations significantly reduces the baseline error rate for the class ``PlayingDaf" with ``PlayingFlute" ($26.8\% \rightarrow 4.9\%$). 
\ifmarked
\red{See \figref{vis} (c).}
\else
See \figref{vis} (c).
\fi
For further details such as confusion matrices, please refer to supplementary material.

\subsection{Experimental results on contrastive learning-based SSL}
\label{sec:comp_TCL}
In previous sections, we mainly conduct experiments with the consistency-based semi-supervised learning framework, FixMatch~\citep{sohn2020fixmatch}.
In this section, we further validate the effectiveness of strong video data augmentation using a contrastive learning-based method, TCL~\citep{singh2021semi}.

\subsubsection{Experimental setup}
\label{sec:setup2}

\topic{Dataset and evaluation.}
Following~\citet{singh2021semi}, we conduct experiments on two large scale action recognition benchmarks: Kinetics-400~\citep{kay2017kinetics} and Mini-Something-v2~\citep{goyal2017something}. The Kinetics-400 dataset consists of 240K videos for training and 20K videos for validation across 400 action classes. The Mini-Something-v2 dataset is a subset of the full Something-Something-v2 dataset, containing 81K training videos and 12K testing videos for 87 action classes.
Using the same data splits and random seeds as~\citet{singh2021semi}, we conduct three random trials for each label ratio and report the Top-1 accuracy with the standard deviations.

\topic{Compared methods.}
We compare our method with the supervised baseline, PseudoLabel~\citep{lee2013pseudo}, MeanTeacher~\citep{tarvainen2017mean}, S4L~\citep{zhai2019s4l}, MixMatch~\citep{berthelot2019mixmatch}, FixMatch~\citep{sohn2020fixmatch}, and the vanilla version of TCL~\citep{singh2021semi}.

\topic{Implementation details.}
We implement our method on top of the publicly available official implementation of TCL.\footnote{\url{https://github.com/CVIR/TCL}}
We use the TSM model~\citep{lin2019tsm} with ResNet-18~\citep{he2016deep} as the feature extraction backbone. 
The vanilla version of TCL contains a fast and a slow branch. 
The fast branch samples four frames for unlabeled data, and the slow branch samples eight frames from the video.
An instance-wise and a group-wise contrastive loss are applied to the data constructed by the two branches. 
Since the fast branch is a temporal augmentation, we only add the photometric-geometric and ActorCutMix into TCL. 
With a probability of 50\%, we treat the fast sampling data followed by photometric-geometric augmentations as the strongly-augmented data. 
Otherwise, we applied ActorCutMix to the slow sampling data and treated it as the strongly-augmented example. 
Lastly, we do \emph{not} adopt label smoothing for ActorCutMix since TCL uses contrastive objective instead of pseudo-labeling. 
We use 2 RTX 2080 GPUs to train our models. 
For other training details such as training schedule and hyper-parameters, we follow the TCL paper~\citep{singh2021semi}.

\subsubsection{Ablation study}
\label{sec:ablation2}
Here, we only validate the effectiveness of the photometric-geometric and ActorCutMix operations since the fast pathway in TCL has already imposed a strong augmentation in the temporal axis.
We conduct ablation experiments on the 1\% label split of both the Kinetics-400~\citep{kay2017kinetics} and the Mini-Something-v2~\citep{goyal2017something} datasets using a TSM model with a ResNet-18 backbone. 
We run the experiments using the first random seed provided by~\citet{singh2021semi}. 
As shown in \tabref{abl_TCL}, both photometric-geometric and ActorCutMix augmentations contribute to the performance of both datasets. 
On the Mini-Something-v2 dataset, ActorCutMix significantly improves performance ($+3.34\%$ gain compared to without ActorCutMix) while photometric-geometric augmentation shows a slight improvement of $+0.59\%$ compared to temporal augmentation only (TCL). 
Mini-Something-v2 requires models to rely more on temporal context, compared to Kinetics-400.
Therefore, the improvement on the Mini-Something-v2 implies that by injecting scene invariance, our data augmentation encourages the model to leverage the temporal context more effectively than the baseline without ActorCutMix.

\begin{table}[t]
\centering

\caption{\tb{Ablation studies with TCL framework}
We report the top-1 accuracy on the $1\%$ labeled split with the first random seed provided by \citet{singh2021semi}. We conduct experiments using a TSM model with a ResNet-18 backbone.
}

\resizebox{\linewidth}{!}{

\begin{tabular}{lcc}
\toprule
Augmentation & Mini-Something-v2 & Kinetics-400   \\
\midrule

Temporal only (TCL) & 7.92 & 8.16 \\

+ Photometric-geometric & 8.51 & 9.00 \\

+ ActorCutMix (All) & \textbf{11.85} & \textbf{9.24} \\
\bottomrule
\end{tabular}
}

\label{tab:abl_TCL}
\end{table}

\subsubsection{Comparison with the state of the arts}
\label{sec:comp2}
We compare our strong data augmentation strategy plugged into TCL to existing methods in \tabref{main_sthv2} and \tabref{main_k400}. On the Mini-Something-v2 dataset, our method consistently achieves favorable performance compared with other methods across all label ratios ($1\%$, $5\%$, and $10\%$). The results show that the proposed strong augmentation strategy, including ActorCutMix, could improve the performance of a state-of-the-art semi-supervised learning framework. Mini-Something-v2 requires models to rely more on temporal context than Kinetics-400, UCF-101, and HMDDB-51. Therefore, the improvement on the Mini-Something-v2 implies that by injecting scene invariance, our data augmentation encourages the model to leverage the temporal context more effectively than TCL. On the Kinetics-400 dataset, we observe a similar trend on the Mini-Something-v2 dataset. Our data augmentation improves upon TCL and achieves state-of-the-art performance on the Kinetics-400 dataset.

\begin{table}[ht]
\centering

\caption{\tb{Results on Mini-Something-v2.}
* indicates we run official code to get the results.
The best performance is in \best{bold} and the second best is \second{underlined}.
}

\resizebox{\linewidth}{!}{

\begin{tabular}{lccc}
\toprule
Method &   1\%  & 5\% &  10\%  \\ 
\midrule
Supervised  &   5.98 $\pm$ 0.68   &  17.26 $\pm$ 1.17  &  24.67 $\pm$ 0.68 \\
PseudoLabel~\citep{lee2013pseudo} & 6.46 $\pm$ 0.32 & 18.76 $\pm$ 0.77 & 25.46 $\pm$ 0.45 \\
MeanTeacher~\citep{tarvainen2017mean} & 7.33 $\pm$ 1.13 & 20.23 $\pm$ 1.59 & 30.15 $\pm$ 0.42 \\
S4L~\citep{zhai2019s4l} & 7.18 $\pm$ 0.97 & 18.58 $\pm$ 1.05 & 26.04 $\pm$ 1.89 \\
MixMatch~\citep{berthelot2019mixmatch} & 7.45 $\pm$ 1.01 & 18.63 $\pm$ 0.99 & 25.78 $\pm$ 1.01 \\
FixMatch~\citep{sohn2020fixmatch} & 6.04 $\pm$ 0.44 & 21.67 $\pm$ 0.18 & 33.38 $\pm$ 1.58 \\
* TCL
~\citep{singh2021semi} & \second{7.52 $\pm$ 0.40} & \second{29.07 $\pm$ 0.71} & \second{40.09 $\pm$ 0.91} \\
Ours & \best{9.93 $\pm$ 1.68} & \best{31.85 $\pm$ 1.32} & \best{42.82 $\pm$ 0.37} \\
\bottomrule
\end{tabular}

}

\label{tab:main_sthv2}
\end{table}

\begin{table}[ht]
\centering

\caption{\tb{Results on Kinetics-400.}
* indicates we run official code to get the results.
The best performance is in \best{bold} and the second best is \second{underlined}.
}

\resizebox{\linewidth}{!}{

\begin{tabular}{lcc}
\toprule
Method &   1\%  & 5\% \\ 
\midrule
Supervised  &  6.17$\pm$0.32  &  20.50$\pm$0.23   \\
PseudoLabel~\citep{lee2013pseudo} & 6.32$\pm$0.19  & 20.81$\pm$0.86 \\
MeanTeacher~\citep{tarvainen2017mean} & 6.80$\pm$0.42 & 22.98$\pm$0.43  \\
S4L~\citep{zhai2019s4l} & 6.32$\pm$0.38 & 23.33$\pm$0.89  \\
MixMatch~\citep{berthelot2019mixmatch} & 6.97$\pm$0.48 & 21.89$\pm$0.22  \\
FixMatch~\citep{sohn2020fixmatch} & 6.38$\pm$0.38 & 25.65$\pm$0.28  \\
* TCL
~\citep{singh2021semi} & \second{7.93$\pm$0.20} & \second{29.32$\pm$0.41}  \\
Ours & \best{9.02$\pm$0.31} & \best{31.45$\pm$0.60}  \\
\bottomrule
\end{tabular}

}

\label{tab:main_k400}
\end{table}

\subsection{Experimental results on fully supervised learning}
\label{sec:full_sup}
\ifmarked
\red{
We demonstrate that our strong data augmentation strategy is general by validating the effectiveness of our strong data augmentation under the fully-supervised (\secref{full_label}) and cross-dataset semi-supervised learning (\secref{cross_semi}) settings. }
\else
We demonstrate that our strong data augmentation strategy is general by validating the effectiveness of our strong data augmentation under the fully-supervised (\secref{full_label}) and cross-dataset semi-supervised learning (\secref{cross_semi}) settings.
\fi

\subsubsection{Full-label setting results}
\label{sec:full_label}
\begin{table}[t]
\centering

\caption{\tb{Full-label setting.}
Our strong data augmentation strategies outperform standard weak data augmentation (\ie random horizontal flipping, scaling, cropping) in the fully-supervised training setting.
}

\begin{tabular}{lcc}
\toprule
Augmentation & UCF-101 & HMDB-51   \\
\midrule

Standard & 55.67 & 40.78 \\

Ours & \textbf{68.31} & \textbf{44.51} \\
\bottomrule
\end{tabular}

\label{tab:full_sup}
\end{table}

Our proposed strong data augmentation can further improve supervised training in the full-label setting.
We train an R(2+1)D ResNet-34 model from scratch and replace the standard weak video augmentation operations (\ie random horizontal flip, scaling, cropping) with our strong data augmentation strategy.
As shown in \tabref{full_sup}, our strong augmentation leads to a significant performance boost on both the UCF-101 (55.67\% $\rightarrow$ 68.31\%) and HMDB-51 (40.78\% $\rightarrow$ 44.51\%) datasets.

\subsubsection{Improving Fully-Supervised Learning with Cross-Dataset Semi-Supervised Learning}
\label{sec:cross_semi}
Lastly, we show that our strong data augmentation strategy can be used in a cross-dataset semi-supervised learning setting to improve the fully-supervised models further.
We use the FixMatch framework for these experiments.
We use the entire training set of the UCF-101 or HMDB-51 datasets as the labeled set and the Kinetics-100 dataset as the unlabeled set.
Since there is no overlap between the labeled and unlabeled data, we modify the semi-supervised framework (as shown in \figref{overview}) by replacing the standard/weak augmentation in the labeled branch with our strong data augmentation to make sure we generate diverse samples for both labeled and unlabeled data.

\begin{table*}[ht]
\centering

\caption{\tb{Cross-dataset semi-supervised learning.}
We train R(2+1)D ResNet-34 models from scratch.
}

\begin{tabular}{lcccc}
\toprule
&   Labeled data  & Unlabeled data &  Top-1 acc. (\%)  & Improvement ($\%$)  \\ 
\midrule
Supervised (Standard Augmentation)   &   UCF-101   &   -   &   55.67  & Reference \\
Supervised (Our Strong Augmentation)   &   UCF-101   &   -   &   68.31 & +12.64 \\
Cross-Dataset Semi-Supervised &  UCF-101   &   Kinetics-100   & 70.31 & +14.64 \\
\midrule
Supervised (Standard Augmentation)   &   HMDB-51   &   -   &   40.78 & Reference \\
Supervised (Our Strong Augmentation)   &   HMDB-51   &   -   &   44.51 & +3.73 \\
Cross-Dataset Semi-Supervised &  HMDB-51   &   Kinetics-100   &  45.75 & +4.97 \\
\bottomrule
\end{tabular}

\label{tab:ssl_cross}
\end{table*}

As shown in \tabref{ssl_cross}, our strong augmentation can be used in a cross-dataset setting (high-data regime), leading to a substantial performance gain compared to the fully supervised training setting with standard/weak data augmentations.
\section{Conclusions}
\label{sec:conclusions}
In this paper, we investigate different types of data augmentation strategies for
video action recognition, in both low-label and full-label settings. 
Our study shows the importance of (1) temporally-coherent photometric and geometric augmentations, (2) temporal augmentations, and (3) actor/scene augmentation.
We show promising action recognition performance on the public benchmarks in both low-label and full-label settings with all the augmentations.
We believe that our exploration help facilitate future video action recognition research.

\\

\heading{CRediT authorship contribution statement}\\

\textbf{Yuliang Zou}: Conceptualization, Methodology, Software, Writing - original draft. \textbf{Jinwoo Choi}: Conceptualization, Methodology, Software, Writing - original draft, Supervision. \textbf{Qitong Wang}: Software, Writing - original draft. \textbf{Jia-Bin Huang}: Writing – original draft, Supervision.\\

\heading{Declaration of competing interest}\\

The authors declare that they have no known competing financial interests or personal relationships that could have appeared to influence the work reported in this paper.\\

\heading{Acknowledgment}\\

This work was supported in part by NSF under Grant No. 1755785 and by a grant from Kyung Hee University in 2021 (KHU-20210735) and by the Institute of Information and Communications Technology Planning and Evaluation (IITP) grant funded by the Korea Government (MSIT) (Artificial Intelligence Innovation Hub) under Grant 2021-0-02068.\\

\bibliographystyle{model2-names}
\bibliography{main}

\end{document}